%% file: main.tex
\documentclass[conference]{IEEEtran}
\usepackage{times}

\usepackage[numbers,sort]{natbib}
\usepackage{multicol}
\input{package}

\input{macro}

\hypersetup{
  pdfauthor={},
  pdftitle={\ours: Eliciting VLM Spatial Intelligence for Generalist Embodied Navigation},
  pdfsubject={Robotics: Science and Systems},
  pdfkeywords={vision-and-language navigation; generalist embodied navigation; spatial intelligence; cross-embodiment transfer}
}

\newcommand{\projectpageicon}{%
  \raisebox{-0.13em}{\includegraphics[height=0.88em]{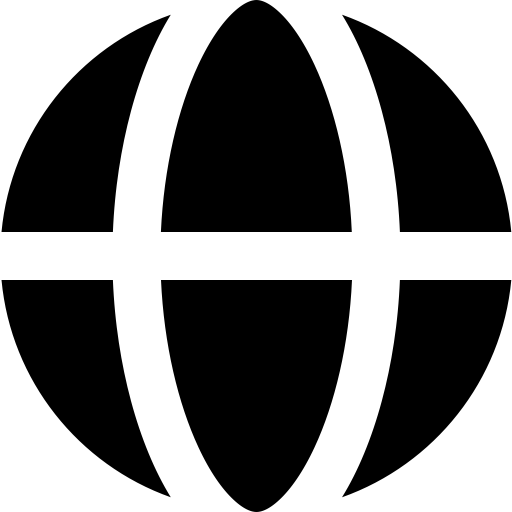}}%
}
\newcommand{\githubicon}{%
  \raisebox{-0.13em}{\includegraphics[height=0.90em]{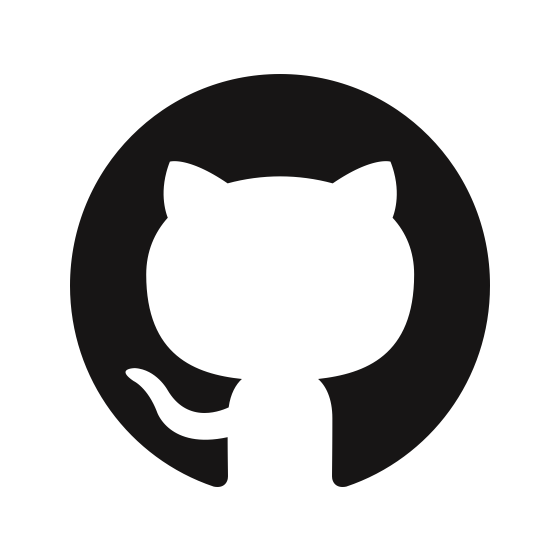}}%
}
\newcommand{\huggingfaceicon}{%
  \raisebox{-0.14em}{\includegraphics[height=0.94em]{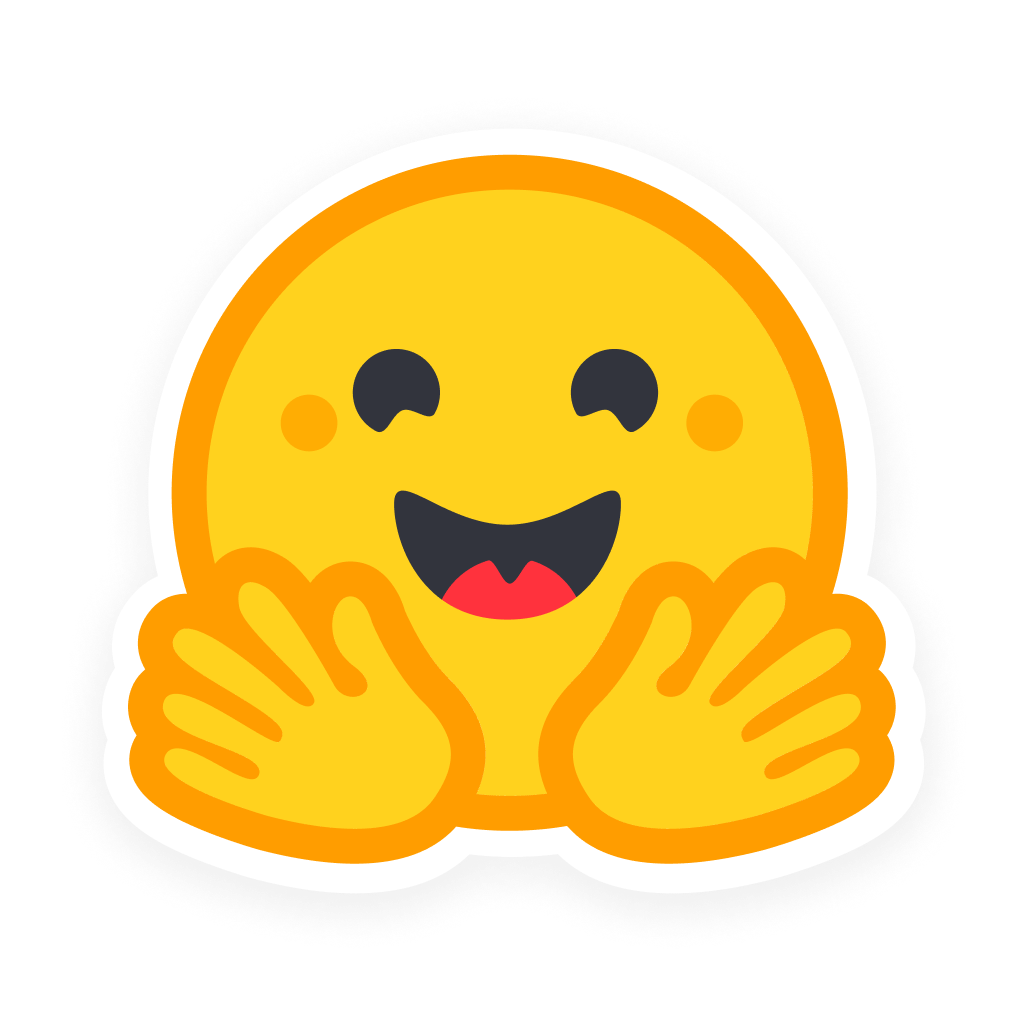}}%
}

\begin{document}

\AddToHookNext{shipout/foreground}{%
  \begin{tikzpicture}[remember picture,overlay]
    \node[anchor=north west,inner sep=0pt]
      at ([xshift=0.68in,yshift=-0.28in]current page.north west) {%
        \includegraphics[
          width=1.55in,
          trim=99.8bp 408bp 83.5bp 377.3bp,
          clip
        ]{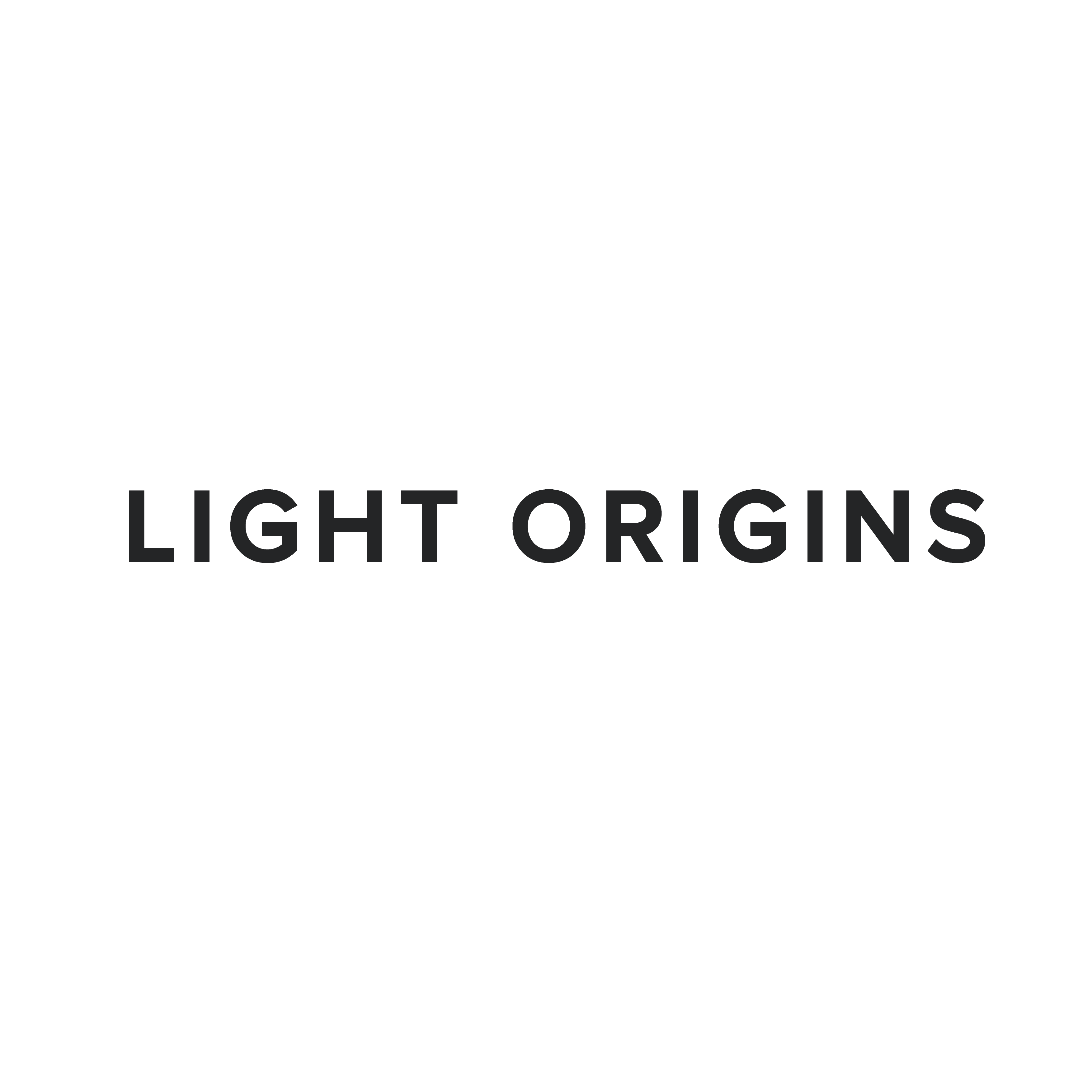}%
      };
    \draw[black!45,line width=0.45pt]
      ([xshift=0.68in,yshift=-0.50in]current page.north west) --
      ([xshift=-0.68in,yshift=-0.50in]current page.north east);
  \end{tikzpicture}%
}

\title{\ours: Eliciting VLM Spatial Intelligence \\ for Generalist Embodied Navigation}

\author{\textbf{Light Origins Team}\\ \\
\small
\href{https://www.lightorigins.com/en/blog/lightnav-0}{\projectpageicon\hspace{0.22em}Project Page}\quad
\href{https://github.com/lightorigins/LightNav-0}{\githubicon\hspace{0.22em}GitHub}\quad
\href{https://huggingface.co/LightOriginsHQ/LightNav-0}{\huggingfaceicon\hspace{0.22em}Hugging Face}}

\maketitle
\vspace{-5em}

\setlength{\stripsep}{2pt}
\begin{strip}
    \vspace*{-2em}
    \centering
    \includegraphics[width=\textwidth]{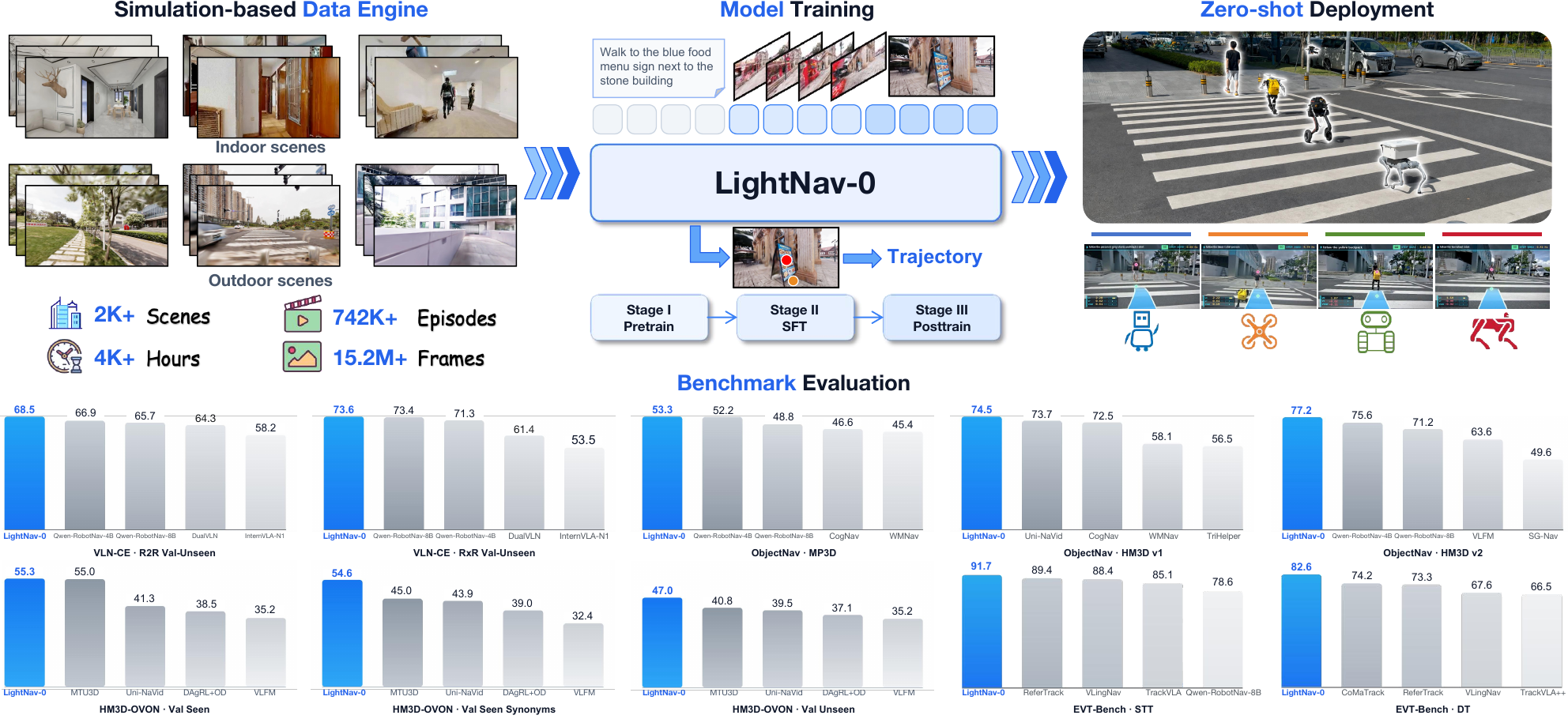}
    \captionof{figure}{\textbf{\ours: a compact generalist embodied navigation model.} A single autoregressive model expresses spatial intent through affordance and object points and predicts RVQ-tokenized trajectories without task- or embodiment-specific prediction heads. The single model achieves state-of-the-art monocular success rates across 10 public simulation settings spanning instruction following, object goal navigation, and embodied visual tracking, and transfers zero-shot to humanoid, quadruped, aerial, and wheeled robots. Blue bars denote \ours, whereas gray bars denote prior methods.}
    \label{fig:teaser}
\end{strip}
\setlength{\stripsep}{15pt plus 2pt minus 2pt}

\begin{abstract}
Embodied navigation requires agents to translate heterogeneous goals and visual observations into actions across tasks, environments, and robot embodiments. Modern vision-language models (VLMs) already encode spatial priors for visual grounding, spatial reasoning, and pointing, but these capabilities are rarely elicited directly for robot control. Existing navigation systems instead rely on task- or embodiment-specific components, fragmenting perception, reasoning, and action while offering limited generalization. Here we present \ours, a compact generalist embodied navigation model that elicits the spatial intelligence of a pretrained VLM and aligns it with navigation, without task-specific prediction heads. \ours represents diverse navigation tasks through a unified token interface: dual-channel pointing expresses task-, scene-, and embodiment-agnostic spatial intent, while a residual vector-quantized action tokenizer maps this intent to precise, embodiment-specific trajectories. Together with temporally aware visual history compression, ER mid-training, supervised fine-tuning, and reinforcement learning, this formulation supports instruction following, open-vocabulary object navigation, and visual tracking within a single model. The navigation training corpus spans $2\mathrm{K}{+}$ scenes and $4\mathrm{K}{+}$ hours of embodied navigation data. LightNav-ER, the embodied-reasoning checkpoint used to initialize \ours, attains the highest complete-set average across 8 embodied-reasoning benchmarks, while \ours achieves state-of-the-art monocular success rates across all 10 public navigation simulation settings. Real-world evaluations further demonstrate zero-shot generalization across robot embodiments, diverse scenes, and static and dynamic targets. These results establish compact VLMs as a unified and transferable backbone for generalist embodied navigation.
\end{abstract}

\IEEEpeerreviewmaketitle

\input{section/intro}
\input{section/related_work}
\input{section/method}
\input{section/dataset}
\input{section/training_recipe}
\input{section/experiments}
\input{section/conclusion}

\section*{Contributors and Acknowledgements}

\medskip
\noindent{\large\bfseries Contributors}

\medskip

\noindent\textbf{Training Infrastructure:} Qianli Ma, Ran Mei, and Jia Wei

\noindent\textbf{Data Infrastructure:} Fei Huang, Shaoan Wang, Jingyi Xu, Yueyu Wang, and Aocheng Luo

\noindent\textbf{Mid-training:} Shaoan Wang and Fan Yang

\noindent\textbf{Supervised Fine-tuning:} Shaoan Wang and Aocheng Luo

\noindent\textbf{Post-training:} Aocheng Luo and Shaoan Wang

\noindent\textbf{Benchmark:} Fei Huang, Shaoan Wang, Jingyi Xu, and Yueyu Wang

\noindent\textbf{Real-world Deployment:} Xiaoyang Wang, Jiangpeng Hu, Xuhao Liu, Hongming Chen, Yuanbin Shao, Yiyang Lin, and Ziliang Li

\noindent\textbf{Writing:} Shaoan Wang, Tingxiang Fan, Aocheng Luo, Fei Huang, Liang Pan, Xinhang Liu, and Yuntao Ma

\noindent\textbf{Project Leads:} Tingxiang Fan and Shaoan Wang

\medskip
\noindent{\large\bfseries Acknowledgements}

\medskip

We thank Bo Liang, Yuxuan Xie, Jiaxin Li, and Tianwei Zhang for their contributions to the early-stage infrastructure and initial technical exploration. We also thank Shiyao Zhang and Shuqi Liao for filming and editing the real-world videos, and Kaisong Chen for designing the cover. We thank our collaborators at LimX Dynamics and Manycore Tech for their support across physical deployment and simulation.

\bibliographystyle{plainnat}
\bibliography{references}

\end{document}

%% file: package.tex
\usepackage{amsmath}
\usepackage{amssymb}
\usepackage{mathtools}
\usepackage{bm}
\usepackage{dsfont}
\usepackage{stmaryrd}
\usepackage{systeme}
\usepackage{scalerel}

\DeclareSymbolFont{timesdigits}{OT1}{ptm}{m}{n}
\SetSymbolFont{timesdigits}{bold}{OT1}{ptm}{b}{n}
\DeclareMathSymbol{0}{\mathalpha}{timesdigits}{`0}
\DeclareMathSymbol{1}{\mathalpha}{timesdigits}{`1}
\DeclareMathSymbol{2}{\mathalpha}{timesdigits}{`2}
\DeclareMathSymbol{3}{\mathalpha}{timesdigits}{`3}
\DeclareMathSymbol{4}{\mathalpha}{timesdigits}{`4}
\DeclareMathSymbol{5}{\mathalpha}{timesdigits}{`5}
\DeclareMathSymbol{6}{\mathalpha}{timesdigits}{`6}
\DeclareMathSymbol{7}{\mathalpha}{timesdigits}{`7}
\DeclareMathSymbol{8}{\mathalpha}{timesdigits}{`8}
\DeclareMathSymbol{9}{\mathalpha}{timesdigits}{`9}
\SetMathAlphabet{\mathrm}{normal}{OT1}{ptm}{m}{n}
\SetMathAlphabet{\mathrm}{bold}{OT1}{ptm}{b}{n}

\usepackage[table]{xcolor}
\usepackage{graphicx}
\usepackage{tikz}
\usetikzlibrary{tikzmark, calc}
\usepackage{float}
\usepackage{stfloats}
\usepackage{placeins}

\usepackage{booktabs}
\usepackage{multirow}
\usepackage{multicol}
\usepackage{makecell}
\usepackage{adjustbox}

\let\labelindent\relax
\usepackage{enumitem}
\usepackage{paralist}
\usepackage{outline}
\usepackage{lipsum}
\usepackage{soul}
\usepackage[normalem]{ulem}
\usepackage{siunitx}
\usepackage{xspace}
\usepackage{url}
\usepackage{cite}

\usepackage{algpseudocode}
\usepackage{listings}

\usepackage[font={small}]{caption} 
\usepackage{subcaption} 
\renewcommand{\thetable}{\Roman{table}}

\usepackage{titletoc}
\usepackage{tocloft}
\usepackage[nolist]{acronym}

\usepackage{balance}

\usepackage{cuted}
\usepackage{esvect}
\usepackage{stackengine}

\usepackage{tcolorbox}
\tcbuselibrary{skins, breakable} 

\newtcolorbox{yellowbox}{
    colback=yellow!30,
    colframe=white,
    boxrule=0pt,
    arc=0pt, 
    left=5pt, right=5pt, top=5pt, bottom=5pt,
    breakable, 
}

\definecolor{citecolor}{HTML}{0071bc}
\usepackage[pagebackref=true,breaklinks=true,urlcolor=blue,colorlinks,citecolor=citecolor,bookmarks=true]{hyperref}
\renewcommand{\backref}[1]{}
\renewcommand{\backrefalt}[4]{}

%% file: macro.tex
\def\ours{\mbox{LightNav-0}\xspace}

\definecolor{myblue}{HTML}{dbe8f5}
\definecolor{mygreen}{HTML}{009900}

\definecolor{BestColor}{HTML}{ffffff}
\definecolor{SecondColor}{HTML}{ffffff}

%% file: section/intro.tex
\section{Introduction}

Embodied navigation is the capability that enables an agent to move through the physical world and reach a location appropriate for accomplishing a given goal. It encompasses a range of goal-directed behaviors, including following natural-language instructions, searching for specified objects or places, and tracking moving targets~\citep{anderson2018vision,yokoyama2024hm3d,wang2025trackvla}. Across these tasks, the agent must ground linguistic or visual goals in its observations, maintain spatial and temporal context, and translate multimodal understanding into executable actions. A general navigation model should therefore transfer not only across environments, but also across tasks and robot embodiments. Most existing systems, however, are optimized for a single task or benchmark and rely on specialized components such as waypoint predictors, topological maps, or action heads~\citep{krantz2021waypoint,an2024etpnav,wang2023gridmm,wang2025trackvla}. This fragmentation limits open-vocabulary transfer, hinders the reuse of learned reasoning across platforms, and isolates embodied navigation from the scaling benefits of modern vision-language models (VLMs).

Modern VLMs already encode many capabilities required for navigation, including open-vocabulary recognition, spatial reasoning, instruction understanding, and temporal video interpretation~\citep{team2023gemini,Hong2024CogVLM2VL,bai2025qwen3vl}. This suggests a different design principle: a compact VLM can serve as a shared reasoning backbone for general embodied navigation. We instantiate this principle with Qwen3-VL-4B-Instruct~\citep{bai2025qwen3vl}, retaining its pretrained architecture without introducing task-specific prediction heads for individual tasks or embodiments. Navigation capabilities are acquired through vocabulary extension, unified supervision, and staged training, keeping the model compact while preserving its semantic and spatial priors.

Bridging a general-purpose VLM and heterogeneous embodied navigation policies requires an intermediate representation that is both spatially meaningful and independent of any particular action space. We argue that pointing naturally provides such an interface. Recent VLMs can express grounded spatial decisions directly as image-plane points~\citep{cheng2025pointarena,yuan2024robopoint}, allowing a navigation model to preserve and exploit the backbone's pretrained capabilities in visual grounding, spatial reasoning, and scene understanding. We therefore formulate dual-channel pointing with channel-specific image-grid tokens as a shared interface across tasks and robot embodiments. An affordance point indicates a feasible direction or free-space waypoint, whereas an object point localizes the task goal, either a target object or a goal location. Predicting this grounded spatial intent before action decoding provides an explicit reasoning step that guides the generation of precise, embodiment-specific navigation actions. This shared representation consequently supports instruction following, object search, and target tracking without separate task-specific designs.

Building on this interface, we construct an end-to-end system using only the compact VLM backbone and token-based outputs. An automatic dual-channel annotation pipeline projects navigation targets into the shared pointing space. Temporally aware visual history compression preserves both recent detail and long-horizon context within a bounded visual-token budget. An RVQ-based action tokenizer converts short-horizon trajectories into language-model tokens, which are subsequently mapped by an execution layer to platform-specific controls. Embodied-reasoning mid-training, supervised fine-tuning with DAgger~\citep{ross2011dagger}, and online reinforcement learning progressively align perception, reasoning, and control.

We evaluate \ours across 10 public navigation simulation settings covering instruction-following VLN, object goal navigation, and embodied visual tracking. Complementary pointing and spatial-VQA evaluations test whether navigation adaptation preserves the VLM's original grounding abilities. Real-world demonstrations further probe transfer across robot embodiments, outdoor scenes, and dynamic targets. Together, these experiments demonstrate a central hypothesis: a compact VLM backbone can provide a unified, transferable basis for cross-task and cross-embodiment navigation.

%% file: section/related_work.tex
\section{Related Work}
\label{sec:related_work}

\subsection{Generalist Embodied Navigation}

Embodied navigation has long been studied as separate problems, including
instruction-following VLN~\citep{anderson2018vision,ku2020room,Krantz2020BeyondTN}, object navigation~\citep{batra2020objectnav,yokoyama2024hm3d}, and embodied visual
tracking~\citep{wang2025trackvla}. Solutions either chain perception, mapping,
and planning modules through hand-designed
interfaces~\citep{yokoyama2024vlfm,yin2024sg,long2024instructnav,yin2025unigoal}
or train a task-specific end-to-end
policy~\citep{ramrakhya2022habitat,zeng2024poliformer}, and both transfer poorly
once the task, the sensor suite, or the robot changes.

Video-based vision-language-action (VLA) models replaced these pipelines with a
single vision-language backbone. NaVid~\citep{zhang2024navid} first showed that a
video VLM can predict navigation actions from monocular RGB alone, and
subsequent models unified more tasks and improved streaming
efficiency~\citep{zhang2024uni,cheng2024navila,wei2025streamvln}. Recent
navigation foundation models further scale this recipe across tasks, environments, and
embodiments, including NavFoM~\citep{zhang2025embodied},
ABot-N0/N1~\citep{chu2026abotn0,gong2026abotn1},
Qwen-RobotNav~\citep{zhang2026qwenrobotnav}, and
Qwen-VLA~\citep{wang2026qwenvla}. Their action interfaces, however, differ
sharply. Discrete atomic commands decoded as language
tokens~\citep{zhang2024navid,wei2025streamvln,qi2025vln} quantize motion
coarsely. Waypoint predictors on panoramic observations or topological
graphs~\citep{krantz2021waypoint,an2024etpnav,wang2023gridmm,wang2024lookahead}
follow instructions well, but demand extra sensing and an explicit map.
Continuous action modules built on anchor-based
diffusion~\citep{wang2025trackvla}, flow
matching~\citep{black2024pi_0,black2025pi_}, or regression
heads~\citep{kim2025fine,zhang2026qwenrobotnav} act smoothly, yet they place
action generation outside the language model. Purely autoregressive
quantization~\citep{kim2024openvla,qu2025spatialvla} keeps generation inside the
model, but trades precision for token efficiency.

Across these designs, transferable navigation capability often relies on structure added
outside the pretrained backbone: panoramic or multi-camera front ends, task and
embodiment identifier tokens, and separate action heads or experts. \ours keeps a
single compact backbone driven by monocular RGB, compresses its observation
history under a fixed visual-token budget, and introduces neither task identifier nor
auxiliary prediction head. Its residual vector-quantized tokenizer decodes 10
$\mathrm{SE}(2)$ waypoints from three tokens of the original language-model head,
enabling high-precision continuous control without a diffusion planner, flow-matching
expert, or embodiment-specific action head.

\subsection{Linguistic and Visual Chain-of-Thought}

Chain-of-thought prompting~\citep{wei2022chain} has been extended to embodied action by producing an explicit reasoning trace before acting. On the linguistic side, embodied chain-of-thought generates grounded language plans before low-level commands~\citep{zawalski2024ECoT,zhouchatvla}. In navigation, OctoNav~\citep{gao2025octonav} and Nav-R1~\citep{liu2025nav} cold-start think-before-action behavior from synthesized chains of thought, and
VLingNav~\citep{wang2026vlingnav} triggers reasoning adaptively rather than at a
fixed rate. Hydra-Nav~\citep{wang2026hydranav} unifies slow temporal-spatial deliberation and fast reactive execution within a single VLM, learning to invoke the slow system selectively at critical navigation stagnation points. Aux-Think~\citep{wang2025think} reports that textual reasoning helps
most as an auxiliary training signal, because decoding long rationales at every
step is costly and can degrade control.

A second family expresses the trace visually rather than in words. CoT-VLA~\citep{zhao2025cot} reasons through predicted future frames, ThinkAct~\citep{huang2025thinkact} reinforces latent visual plans, and NavForesee~\citep{liu2025navforesee} plans hierarchically inside a vision-language world model. Several methods place the trace on the image plane: AO-Planner~\citep{chen2024affordances} selects affordance-grounded pixel
waypoints with a VLM and delegates execution to a low-level planner. The
dual-system models DualVLN and
InternVLA-N1~\citep{wei2025dualvln,internvla2025} ground a farthest reachable
pixel goal with a slow 7B planner, then hand it to a fast diffusion-style
trajectory policy. Robostral Navigate~\citep{bounhar2026robostral} keeps the
trace inside one monocular model, predicting the next waypoint by pointing in the
current camera view. Others make the trace geometric instead of
positional, using priors from a 3D geometry foundation model~\citep{zeng2025janusvln} or a single occupancy token supervised by volumetric prediction and re-injected as a spatial chain of thought~\citep{liu2026spannav}.

\ours combines explicit reasoning with visual grounding in a fixed-length trace. Its dual-channel pointing prefix predicts an affordance point for feasible local motion and an object point for a target object or goal location. Both are represented with channel-specific image-grid tokens that leverage the
backbone's pretrained spatial grounding rather than replacing it, and the same
representation supports
instruction-following VLN, open-vocabulary object navigation, and embodied
visual tracking. The prefix is supervised in the
same token space as the action and costs a constant number of tokens per
decision. It therefore keeps the interpretability of an explicit reasoning step,
without the variable decoding latency of free-form textual deliberation or a
second planning system.

\subsection{Reinforcement Learning Post-Training}

Reinforcement learning has long complemented imitation in embodied navigation,
from large-scale distributed on-policy training~\citep{wijmans2020ddppo} and its
transformer-scale successors~\citep{zeng2024poliformer} to imitation pretraining
followed by RL fine-tuning~\citep{ramrakhya2023pirlnav}. Those policies are
task-specific and act over a handful of discrete primitives, so an action
probability is immediate and a rollout is cheap. Neither property is automatic once the
policy becomes a generalist VLA.

For VLM-based policies the dominant recipe is verifiable-reward post-training.
Group-relative policy optimization~\citep{shao2024deepseekmath,guo2025deepseek}
removed the learned critic and made a second RL stage practical at scale, and the
same recipe was carried into multimodal
reasoning~\citep{feng2025video,huang2025vision}. Navigation adopted it quickly.
VLN-R1~\citep{qi2025vln} shapes a time-decayed reward over multi-step action
predictions, and Nav-R1~\citep{liu2025nav} combines format, understanding, and
trajectory rewards. OctoNav~\citep{gao2025octonav} refines its reasoning traces
with verifiable rewards before an online exploration stage, and
ActiveVLN~\citep{zhang2025activevln} extends optimization to multi-turn
on-policy rollouts. ABot-N1~\citep{gong2026abotn1} moves the same machinery one
level up. It treats the joint chain-of-thought and pixel-goal output of its slow
system as the action, and shapes format, target, and safety-clearance rewards
over that output, though only for its point-goal task. Robostral
Navigate~\citep{bounhar2026robostral} instead reduces the reward to a single
terminal distance to the goal, and optimizes it with CISPO under group-relative
advantage estimation. It also mines its rollout pool, keeping only the tasks
that its supervised policy fails to solve. These systems differ in reward design, yet they agree on the action
interface: what the optimizer sees is always a discrete language token. That choice is
what keeps the update cheap, because the quantity a policy gradient needs is
already a token log-probability. Its cost is a coarse action space.

Policies that attach a continuous head recover precision and lose that quantity.
SimpleVLA-RL~\citep{li2025simplevla} therefore keeps an autoregressive policy and
supplies outcome-only rewards, whereas flow-based policies must first recast
their denoising process as a Markov decision process before policy gradients
apply~\citep{chen2025pi_,zhang2025reinflow}.

The open question is therefore not how to reshape reward, but whether a policy
can expose a tractable per-action probability while still acting precisely. The
residual vector-quantized interface of \ours supplies both at once. The trajectory at
each decision step is three RVQ tokens emitted by the same head that produces
language, so their log-probabilities are exact and group-relative updates apply
unchanged. No auxiliary MDP, denoising reformulation, or separate critic is
required, yet decoding still recovers 10 continuous $\mathrm{SE}(2)$ waypoints.
Because a whole trajectory costs three tokens, the credit-assignment horizon
stays short and each rollout stays cheap.

%% file: section/method.tex
\section{Model Architecture}

\begin{figure*}[!t]
    \centering
    \includegraphics[width=\linewidth]{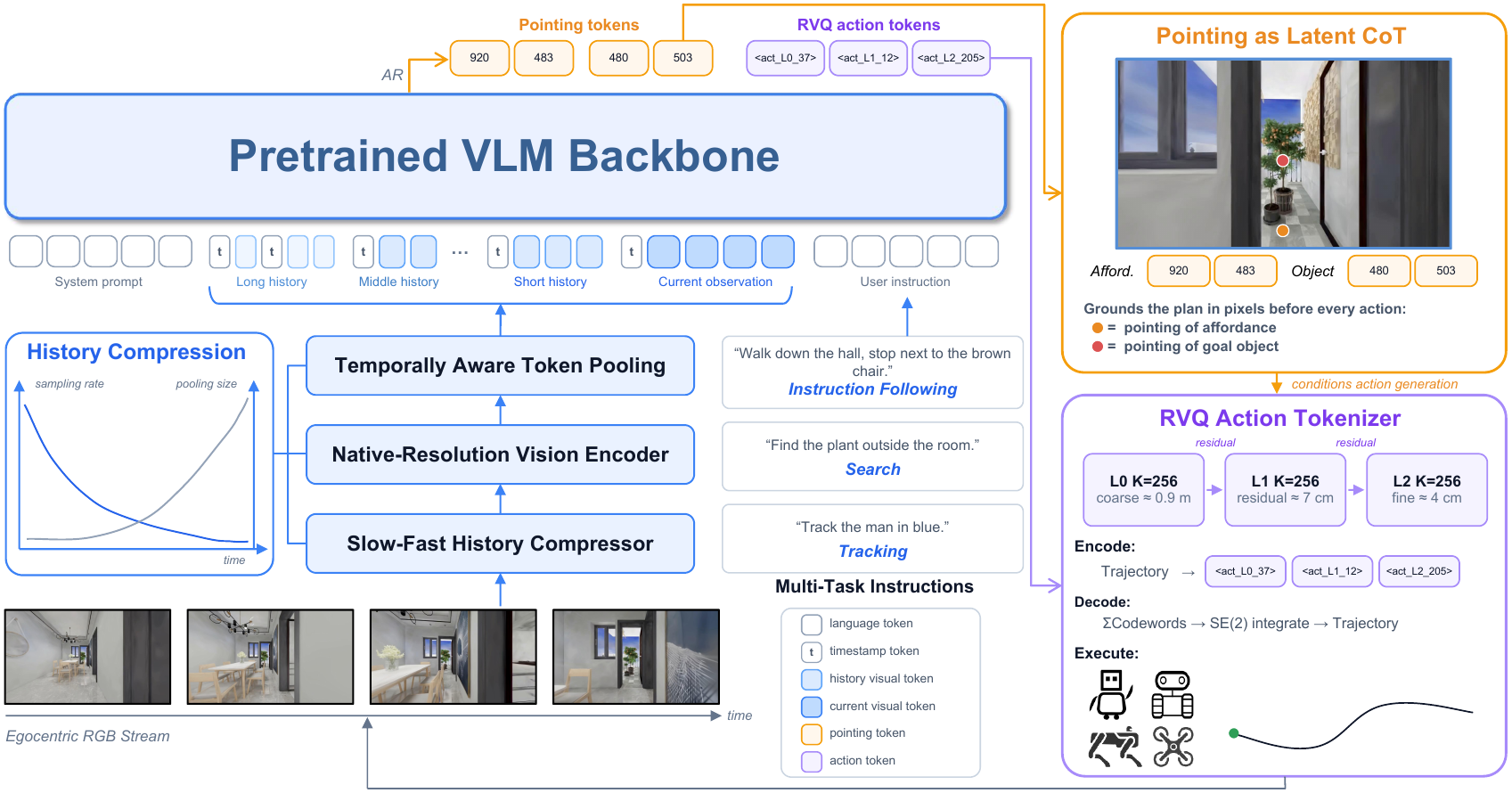}
    \caption{\textbf{Overview of \ours.} A compact pretrained VLM consumes a temporally compressed egocentric RGB history and a natural-language goal. It first emits pointing as an explicit spatial reasoning trace, followed by three residual vector-quantized (RVQ) action tokens. The action tokens decode to 10 future $\mathrm{SE}(2)$ waypoints and are executed by an embodiment-specific low-level controller. The same backbone, token interface, and objective are used for all navigation tasks.}
    \label{fig:model-architecture}
\end{figure*}

\subsection{Architecture Overview}

\ours formulates heterogeneous embodied navigation tasks as conditional token generation. At decision step $t$, the model receives a natural-language instruction $\mathcal{I}$ and an egocentric RGB history $\mathcal{O}_{1:t}=\{\mathbf{o}_1,\ldots,\mathbf{o}_t\}$. It generates a dual-channel pointing prefix followed by a short-horizon action sequence. The decoded action is a trajectory of 10 future $\mathrm{SE}(2)$ waypoints, which provides a common geometric interface to the low-level controllers of different robot embodiments. Task semantics are specified entirely by the instruction and supervision format; we introduce no task-identification token.

\begin{figure}[!t]
    \centering
    \includegraphics[width=0.9\columnwidth,pagebox=cropbox]{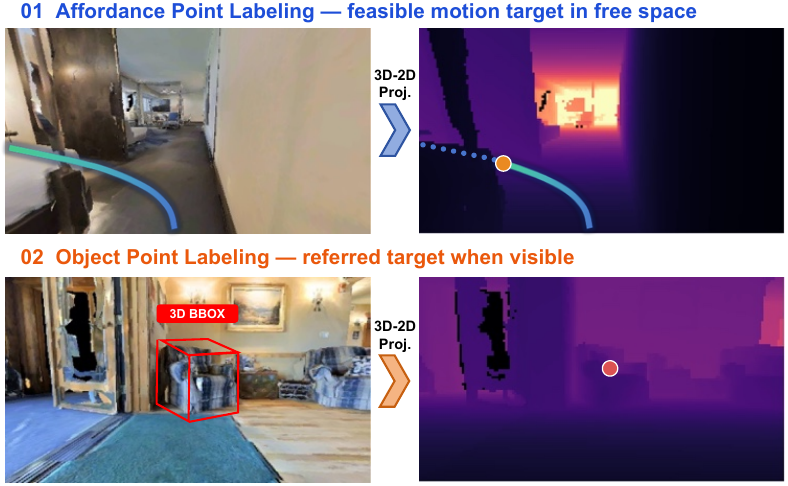}
    \caption{\textbf{Automatic dual-channel pointing annotation.} The affordance channel indicates a feasible local direction or free-space waypoint, whereas the object channel localizes the task goal, either a target object or a goal location.}
    \label{fig:pointing-annotation}
\end{figure}

We instantiate \ours from Qwen3-VL-4B-Instruct~\citep{bai2025qwen3vl}, whose backbone comprises a native-resolution vision transformer and a 36-layer language model. Rather than introducing navigation-specific modules, we retain the pretrained architecture and augment only its vocabulary with indexed $\langle\mathrm{apos}_{i}\rangle$, $\langle\mathrm{opos}_{i}\rangle$, and RVQ action tokens. Both intermediate spatial predictions and action codes are consequently decoded through the original autoregressive language-model head, with no waypoint predictor, task-specific action head, or embodiment-specific expert. As shown in Fig.~\ref{fig:model-architecture}, tokens from the timestamped visual history, current observation, and instruction are interleaved within a single causal sequence at each decision step. This compact formulation preserves the backbone's pretrained visual-grounding and spatial-reasoning capabilities while aligning them with a unified control interface shared across tasks and embodiments.

\begin{figure*}[!t]
    \centering
    \includegraphics[width=\textwidth,pagebox=cropbox]{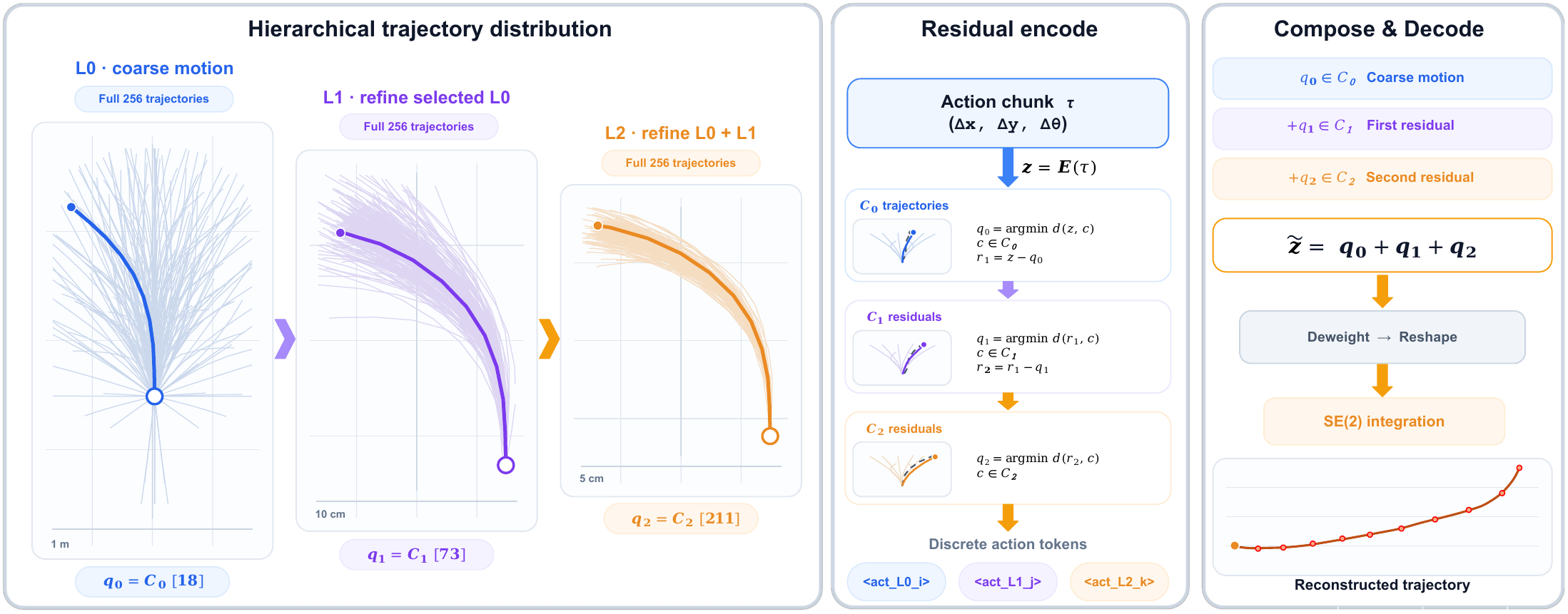}
    \caption{\textbf{Hierarchical residual vector-quantized action tokenizer.} A 10-step $\mathrm{SE}(2)$ trajectory is quantized by a coarse 256-entry codebook and two residual 256-entry codebooks. Each level visualizes all 256 candidates and highlights the selected codeword. Any non-empty token prefix decodes into an executable coarse trajectory, while successive residual levels progressively refine geometric precision.}
    \label{fig:rvq-tokenizer}
\end{figure*}

\subsection{Temporally Aware Visual History Compression}

Navigation requires both recent geometric detail and long-horizon context, but encoding every historical frame at native resolution causes the visual-token count to grow without bound. We therefore compress history according to temporal recency. The design follows the qualitative form of the Ebbinghaus forgetting curve~\citep{ebbinghaus2013image}: recent observations receive a higher sampling rate and finer spatial resolution, whereas older observations are sampled less frequently and pooled more aggressively.

For a historical frame acquired at time $t_i$, we define its age as $\Delta T_i=t-t_i$. Its sampling rate decays exponentially,
\begin{equation}
    f_s(i)=f_s^{\max}\exp\!\left(-\frac{\Delta T_i}{\tau_s}\right)
    \label{eq:history-sampling}
\end{equation}
where $f_s^{\max}$ is the maximum sampling rate and $\tau_s$ controls the temporal decay. The selected frames are encoded independently by the native-resolution vision transformer. Given the resulting patch grid $\mathbf{V}_i$, we set the spatial pooling stride to
\begin{equation}
    \begin{gathered}
        s_i = \max\!\left\{1,\left\lfloor\exp\!\left(\frac{\Delta T_i}{\tau_p}\right)\right\rfloor\right\} \\[0.35em]
        \widetilde{\mathbf{V}}_i = \mathcal{G}_{s_i}\!\left(\mathbf{V}_i\right)
    \end{gathered}
    \label{eq:history-pooling}
\end{equation}
where $\tau_p$ controls the rate of spatial compression and $\mathcal{G}_{s_i}$ denotes grid pooling with stride $s_i$. Thus, temporally distant observations contribute fewer and coarser tokens, while the current observation retains the finest visual detail. Timestamp tokens preserve the temporal ordering after pooling.

The compressor operates after the vision transformer and supports variable-length, native-resolution inputs under configurable pixel budgets of $256\mathrm{K}$, $576\mathrm{K}$, and $1\mathrm{M}$. Within the selected budget, frame sampling and tier-wise adaptive pooling jointly allocate tokens across long-, middle-, and short-term history. This slow-fast allocation bounds the context length without collapsing the entire history to a single fixed-resolution representation.

\subsection{Dual-Channel Pointing as Latent Spatial Reasoning}
\label{subsec:pointing}

A VLM represents visual evidence on a 2D token lattice, whereas navigation requires geometrically precise actions in metric space. We therefore encode each projected point as one channel-specific image-grid token rather than as separate horizontal and vertical coordinate tokens. Let the current view be partitioned into $H_g$ rows and $W_g$ columns. A projected point $\mathbf{p}=(u,v)\in[0,1]^2$ is assigned a flattened grid index
\begin{equation}
    \begin{aligned}
        r(\mathbf{p}) &= \min\!\left\{H_g-1,\left\lfloor H_gv\right\rfloor\right\} \\[0.35em]
        c(\mathbf{p}) &= \min\!\left\{W_g-1,\left\lfloor W_gu\right\rfloor\right\} \\[0.35em]
        i(\mathbf{p}) &= r(\mathbf{p})W_g+c(\mathbf{p})
    \end{aligned}
    \label{eq:grid-pointing}
\end{equation}
The affordance and object channels use distinct token families indexed over this shared lattice. An affordance point $\mathbf{p}^{a}$ is encoded directly as the single token $\langle\mathrm{apos}_{i^{a}}\rangle$, where $i^{a}=i(\mathbf{p}^{a})$. The token family $\langle\mathrm{apos}_{i}\rangle$ represents affordance points, and the selected cell indicates a feasible local direction or landing location in free space. An object point $\mathbf{p}^{o}$ is encoded directly as $\langle\mathrm{opos}_{i^{o}}\rangle$, where $i^{o}=i(\mathbf{p}^{o})$. The token family $\langle\mathrm{opos}_{i}\rangle$ represents object points and localizes the task goal, either a target object or a goal location. Reserved indices within the corresponding token family represent cases without a valid image-grid cell, including in-place turns, stopping, and target invisibility.

At each decision step, the complete navigation output is serialized as
\begin{equation}
    \begin{aligned}
        \mathbf{y}_t = [&\langle\mathrm{apos}_{i_t^{a}}\rangle,\langle\mathrm{opos}_{i_t^{o}}\rangle, \\[0.35em]
        &\langle\mathrm{act\_L0}_{k_{0,t}}\rangle,\langle\mathrm{act\_L1}_{k_{1,t}}\rangle,\langle\mathrm{act\_L2}_{k_{2,t}}\rangle]
    \end{aligned}
    \label{eq:navigation-token-sequence}
\end{equation}
The pointing prefix therefore contains exactly two atomic tokens, followed immediately by three RVQ action tokens; no separate channel-marker or coordinate token is generated. Causal attention makes this compact prefix an explicit latent spatial trace that conditions trajectory generation. The same grid indexing scheme is shared across navigation, pointing, and spatial-VQA supervision, allowing navigation training to reuse the backbone's visual grounding capability. As illustrated in Fig.~\ref{fig:pointing-annotation}, automatic annotation projects the affordance target and the task-relevant object or goal location into the current view. Each valid affordance projection is assigned an indexed $\langle\mathrm{apos}_{i}\rangle$ token, whereas each valid object or goal projection is assigned an indexed $\langle\mathrm{opos}_{i}\rangle$ token. Because the lattice is defined in the image plane rather than in an embodiment-specific control space, the representation remains common across navigation tasks and robot platforms.

\subsection{Residual Vector-Quantized Action Tokenizer}

Directly emitting continuous controls from a language-model head creates a mismatch between token prediction and geometric precision. We instead represent each action chunk as 10 future $\mathrm{SE}(2)$ waypoints and tokenize the complete trajectory with residual vector quantization, as visualized in Fig.~\ref{fig:rvq-tokenizer}. Let $\mathbf{z}_t\in\mathbb{R}^{10\times3}$ denote the vectorized waypoint sequence. The tokenizer contains 3 level-specific codebooks $\mathcal{C}^{(0)},\mathcal{C}^{(1)},\mathcal{C}^{(2)}$, each with 256 codewords. The first level captures the coarse trajectory, and the next two levels successively quantize its residual:
\begin{align}
    k_{\ell} &= \arg\min_k d_J\!\left(\mathbf{r}^{(\ell)},\mathbf{e}^{(\ell)}_k\right) \\
    \mathbf{r}^{(\ell+1)} &= \mathbf{r}^{(\ell)}-\mathbf{e}^{(\ell)}_{k_{\ell}}
    \qquad \mathbf{r}^{(0)}=\mathbf{z}_t
\end{align}
where $d_J$ is a Jacobian-weighted trajectory distance used during residual $k$-means fitting. Weighting in the integrated trajectory space balances translation and heading errors when forming the codebooks. We assign trajectories using
\begin{equation}
    d_{\mathrm{traj}}(\mathbf{z},\hat{\mathbf{z}})=\operatorname{ADE}(\mathbf{z},\hat{\mathbf{z}})+\lambda\left|\Delta\theta(\mathbf{z})-\Delta\theta(\hat{\mathbf{z}})\right|
    \label{eq:trajectory-distance}
\end{equation}
which measures both positional deviation and accumulated heading error at the trajectory level. Based on the trajectory-error analysis, we set $\lambda=0.3$ throughout all experiments.

The three level-specific indices are emitted as level-specific action tokens $[\langle\mathrm{act\_L0}_{k_{0,t}}\rangle,\langle\mathrm{act\_L1}_{k_{1,t}}\rangle,\langle\mathrm{act\_L2}_{k_{2,t}}\rangle]$. For a generated prefix of length $L$, we reconstruct
\begin{equation}
    \hat{\mathbf{z}}_t^{(L)}=\sum_{\ell=0}^{L-1}\mathbf{e}^{(\ell)}_{k_{\ell}}
    \qquad L\in\{1,2,3\}
    \label{eq:rvq-prefix-decoding}
\end{equation}
and integrate $\hat{\mathbf{z}}_t^{(L)}$ in $\mathrm{SE}(2)$ to recover 10 waypoints. The first token specifies a coarse trajectory, and each additional token corrects the residual left by the preceding levels. Every non-empty prefix is reshaped and integrated by the same decoder, so it yields an executable trajectory rather than an incomplete action representation. Under tight compute or latency budgets, generation can therefore stop after the first RVQ token or after two tokens and trade geometric precision for lower autoregressive cost. Full 3-level decoding provides the highest precision and represents up to $256^3$ code combinations with only 3 tokens. It achieves an average displacement error of $0.72\,\mathrm{cm}$, compared with $2.48\,\mathrm{cm}$ for a single-level $K=4096$ VADv2-style planning vocabulary~\citep{chen2024vadv2}. This baseline follows the vectorized planning formulation introduced by VAD~\citep{related_work_driving_VAD}, but uses a single trajectory token without residual refinement. The shared trajectory is finally converted to platform-specific commands by the robot's low-level controller.

\subsection{Unified Autoregressive Objective}

Navigation and auxiliary VQA examples are trained with the same causal language-model objective. For an input context $\mathbf{x}$ and supervised output positions $\mathcal{M}$, we minimize
\begin{equation}
    \mathcal{L}_{\mathrm{CE}}=-\sum_{j\in\mathcal{M}}\log p_{\theta}\!\left(y_j\mid\mathbf{x},y_{<j}\right).
\end{equation}

For navigation, the supervised sequence contains one $\langle\mathrm{apos}_{i}\rangle$ token, one $\langle\mathrm{opos}_{i}\rangle$ token, and 3 RVQ action tokens; for pointing or spatial VQA, it contains the corresponding indexed token or language response. This formulation exposes every task through a shared token space and prediction head. It also avoids separately balanced navigation losses and permits all samples to share the same packed autoregressive training loop.

\subsection{Training and Inference Efficiency}

The compact, purely autoregressive interface allows variable-length samples to be packed without padding each sample to a common maximum. We dynamically pack approximately 8.6 samples into each 8,192-token training sequence. Fused vision rotary-position-embedding operations and an LM-head dimension aligned to a multiple of 8 further reduce kernel and memory overhead. These optimizations apply uniformly because the model contains no task-specific action modules.

At inference time, the vision transformer and language model run in a single process, and autoregressive decoding is served with vLLM~\citep{kwon2023efficient}. This single-process implementation avoids inter-process transfer of visual features and achieves an inference latency of approximately $4\,\mathrm{ms}$ per generated token on an NVIDIA GeForce RTX 4090. Each inference step requires the $\langle\mathrm{apos}_{i}\rangle$ and $\langle\mathrm{opos}_{i}\rangle$ prefix followed by at most 3 RVQ action tokens. Resource-constrained deployments may stop after the first or second RVQ level and execute the corresponding coarse trajectory, whereas the third level is generated when maximum geometric precision is required.

\begin{figure*}[!t]
    \centering
    \includegraphics[width=\textwidth,pagebox=cropbox]{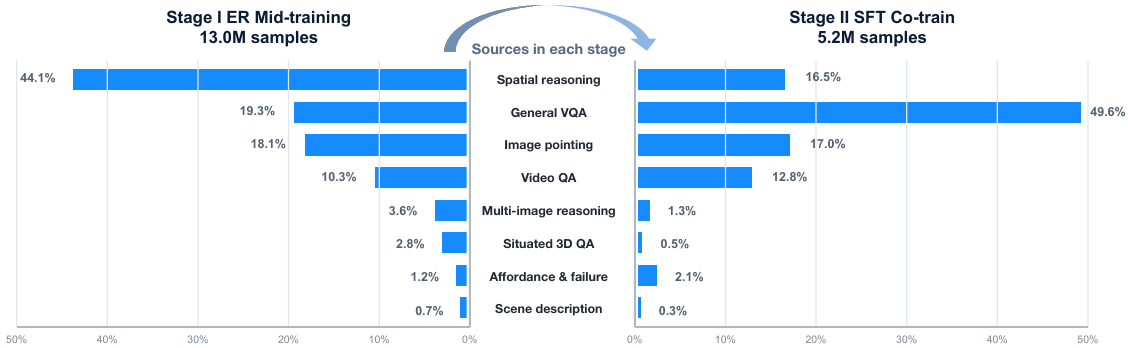}
    \caption{\textbf{2-stage embodied-reasoning data curriculum.} Stage I covers spatial reasoning, general VQA, image pointing, video QA, multi-image reasoning, situated 3D QA, affordance and failure understanding, and scene description. Stage II retains these capabilities through a rebalanced ER mixture during supervised co-training.}
    \label{fig:two-stage-vqa-data}
\end{figure*}

%% file: section/dataset.tex
\section{Data \& Benchmarks}
\label{sec:data-and-benchmarks}

Our data design follows the same principle as the model: heterogeneous embodied tasks should share a common perceptual and spatial basis before they are mapped to actions. We therefore organize training data into two coupled mixtures. The navigation corpus spans $2\mathrm{K}{+}$ scenes and $4\mathrm{K}{+}$ hours of embodied trajectories. Embodied-reasoning (ER) mid-training draws from 36 sources to strengthen spatial understanding, temporal reasoning, and visual grounding. Supervised fine-tuning (SFT) then combines 16 navigation sources with 33 auxiliary ER and VQA sources. Under the active sampling schedule, $77.6\%$ of optimization samples carry navigation-action supervision and $22.4\%$ rehearse the perceptual and reasoning capabilities acquired during ER mid-training. These proportions describe the task-balanced optimization mixture rather than unique corpus coverage. This separation lets us increase task and scene diversity without changing the unified pointing--action interface.

\subsection{Embodied Reasoning Data}

Following the capability-oriented organization of Molmo2-ER~\citep{fang2026molmoact2}, we construct the ER corpus around the competencies that most directly support navigation. The two-stage data curriculum is shown in Fig.~\ref{fig:two-stage-vqa-data}. The mixture allocates $35.14\%$ of its sampling mass to pointing, $25.05\%$ to single-image VQA, $19.81\%$ to video reasoning, and $20.00\%$ to general visual and abstract reasoning. Rather than optimizing for any single benchmark format, these sources expose the model to complementary supervision across synthetic scenes, real images, egocentric video, multi-view observations, and robot interaction data.

\subsubsection{Image Embodied QA}

Eleven single-image VQA sources provide indexed supervision. Three standalone collections are SenseNova-SI Spatial, CLEVR Spatial VQA, and SAT Spatial VQA. The remaining eight are VSTP-SI Depth Comparison, VSTP-SI Distance, VSTP-SI Scene Caption, VSTP-SI Measurement, VSTP-MI Correspondence, VSTP-MI Object--Object Relation, VSTP-MI Camera Motion, and VSTP-MI Scene Caption. Their questions cover relative position, metric depth and distance, object--object relations, camera motion, physical measurement, scene description, and affordance-oriented reasoning. This mixture teaches the model to recover both qualitative relations, such as left/right and in front/behind, and quantitative cues needed to distinguish traversable free space from visually plausible but geometrically invalid targets.

\subsubsection{Video Embodied QA}

Nine video sources provide indexed supervision. Robot-centric supervision comes from RoboVQA-Reasoning, RoboVQA-Understanding, and RoboFAC Failure-VQA. Broader video and situated-spatial supervision comes from VSI-590K Spatial, SIMS-VSI Spatial, ViCA-322K, LLaVA-Video-VQA, SQA3D-Situated, and SpatialLadder Spatial. The data span short robot interactions and clips of up to 64 frames, with supervision for temporal ordering, trajectory-aware spatial relations, planning, affordance prediction, future-state reasoning, and failure understanding. Video supervision is important for navigation because action feasibility depends not only on the current view but also on how objects, people, and the agent itself evolve over time.

\subsubsection{Pointing and Grounding}

Pointing is the largest specialized component of ER mid-training: 13 sources account for $35.14\%$ of the sampling probability. Object and referring-point supervision is drawn from RefSpatial-2D and RefSpatial-3D~\citep{zhou2025roborefer}, PixMo Points Single, PixMo Points Multi, COCO Pointing~\citep{lin2014microsoft}, CoSyn Point, RefL4, and RoboRefIt. Affordance, free-space, and trajectory-point supervision comes from RoboPoint~\citep{yuan2024robopoint}, RoboAfford, HANDAL, FSD Free-Point, and FSD Visual-Trace. Together, these sources cover referred-object localization, free-space selection, interaction affordances, and visual trajectory traces. They teach the VLM to express spatial decisions directly in the image plane. During navigation SFT stage, the same capability is instantiated as the grid used by the affordance-point token $\langle\mathrm{apos}_{i}\rangle$ and object-point token $\langle\mathrm{opos}_{i}\rangle$.

\subsubsection{Multi-image and Ego--Exo Correspondence}

Multi-image samples are drawn from both the image-VQA and video pools. The explicitly multi-image subset comprises VSTP-MI Correspondence, VSTP-MI Object--Object Relation, VSTP-MI Camera Motion, and VSTP-MI Scene Caption, while temporal cross-view samples are inherited from the 9 video sources listed above. They require the model to associate objects across viewpoints, reconcile egocentric and exocentric observations, estimate camera motion, and preserve spatial relations when the visual frame changes. This supervision complements history compression: although the navigation policy receives a temporally compressed context, it must still recognize that observations captured at different times or viewpoints refer to the same scene structure.

\subsubsection{Abstract Embodied Reasoning}

3 general sources account for $20.00\%$ of the ER sampling mass: LLaVA-OneVision Spatial VQA, Euclid30K-Math, and MMIF-23K-Instruct. They cover broad visual question answering, instruction understanding, mathematical reasoning, and compositional spatial relations. We retain this component to prevent specialization from collapsing the linguistic and visual breadth inherited from the pretrained VLM. Synthetic relation problems additionally isolate frame-of-reference and multi-step composition from the appearance biases of natural images.

\subsection{Navigation Data}

\begin{figure*}[!t]
    \centering
    \includegraphics[width=\textwidth,pagebox=cropbox]{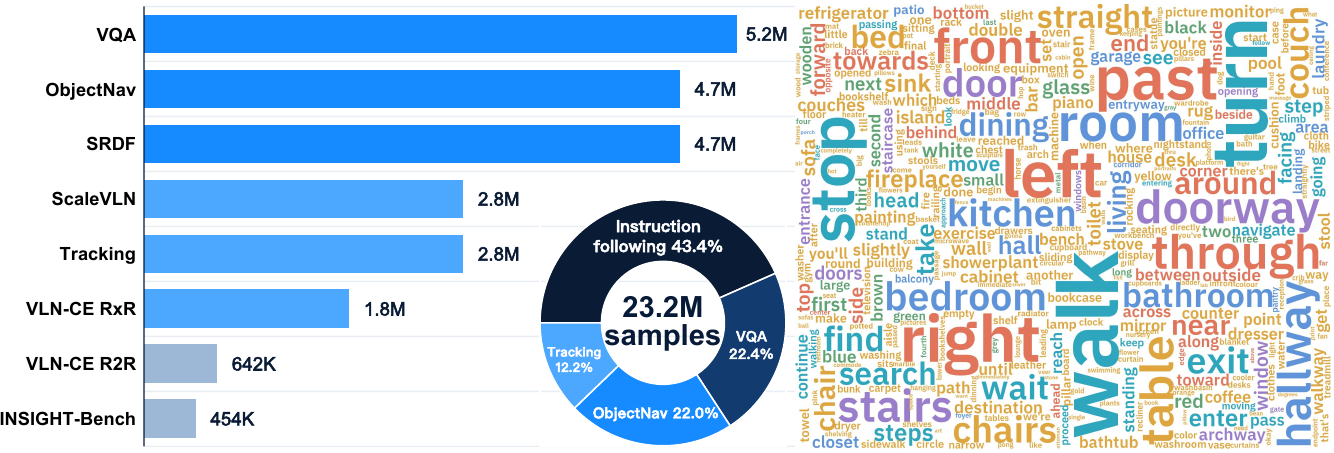}
    \caption{\textbf{Composition of the supervised fine-tuning corpus.} The task-balanced optimization mixture combines VQA with instruction following, object navigation, tracking, VLN-CE, ScaleVLN~\citep{wang2023scalevln}, SRDF~\citep{wang2024srdf}, and INSIGHT-Bench data. The task proportions and instruction word cloud illustrate the diversity of the unified training corpus.}
    \label{fig:sft-data-composition}
\end{figure*}

Navigation supervision is organized by task objective rather than by robot platform. It spans instruction following, object goal navigation, and embodied visual tracking. The task-level composition of the supervised mixture is summarized in Fig.~\ref{fig:sft-data-composition}. To reduce dependence on a fixed sensor configuration, we apply camera randomization throughout navigation-data generation. The field of view, camera height, and pitch are sampled from $[90^\circ,130^\circ]$, $[0.5,1.5]\,\mathrm{m}$, and $[-15^\circ,15^\circ]$, respectively. This augmentation broadens the visual geometries and viewpoints represented by the training trajectories. All samples are converted to the same output sequence: an affordance point, an object point, and three residual vector-quantized trajectory tokens. Consequently, instruction following, object search, and target following can be interleaved in a single autoregressive training stream even though they differ in goal semantics, temporal structure, and termination conditions.

\subsubsection{Instruction Following}

The instruction-following mixture combines randomized-camera expert trajectories from R2R~\citep{anderson2018vision} and RxR~\citep{ku2020room} with self-distilled routes and ScaleVLN data. The synthesized and self-distilled sources provide substantially broader scene and instruction diversity than the human-annotated corpora alone. Together, expert, synthesized, and self-distilled routes expose the model to both fine-grained linguistic alignment and large-scale geometric variation.

\subsubsection{Object-Goal Navigation}

The object-goal mixture contains semantic and expert demonstrations derived from PIRLNav~\citep{ramrakhya2023pirlnav}, HM3D~\citep{ramakrishnan2021hm3d}, MP3D~\citep{chang2017matterport3d}, VLNVerse~\citep{lin2025vlnverse}, Habitat-GS~\citep{xia2026habitatgs}, and InteriorGS~\citep{miao2026towards}. We further self-collect exploration trajectories from HM3D-OVON~\citep{yokoyama2024hm3d} and MP3D, jointly supervising a feasible landing region and the referred object. Finally, in-loop DAgger samples expose the policy to states induced by its own actions rather than only states visited by an expert.

\subsubsection{Embodied Visual Tracking}

Tracking data contain randomized-camera person-following trajectories derived from EVT-Bench~\citep{wang2025trackvla}. Unlike goal-reaching tasks, tracking requires persistent target identity and continuous relative-position control; including it in the same mixture therefore broadens the temporal behavior learned by the shared policy.

\subsubsection{Quality Control and Sampling}

We apply task-aware filters before constructing the balanced training index. For instruction following and object navigation, stop supervision is retained only at the final frame and only when the target is visible; episodes in which the target is never observed are removed. Stop samples are capped at $2\%$ of the mixture to prevent premature termination from dominating token prediction. Dual-channel pointing labels use a shared grid format, while each 10-waypoint action chunk is encoded by three $K=256$ RVQ levels. To reduce domination by common motion patterns, trajectory clusters are balanced with a maximum cluster share of $5\%$.

\subsection{Evaluation Benchmarks}

We evaluate the ER checkpoint on 8 benchmarks that isolate complementary spatial capabilities: Point-Bench~\citep{cheng2025pointarena}, RefSpatial~\citep{zhou2025roborefer}, the POI and VQA tracks of RoboSpatial~\citep{song2024robospatial}, Where2Place~\citep{yuan2024robopoint}, CV-Bench~\citep{tong2024cambrian}, ERQA~\citep{geminirobotics2025}, and EmbSpatial~\citep{du2024embspatial}. Together they measure visual pointing, spatial referring, interaction-site prediction, affordance grounding, geometric perception, and embodied question answering.

Navigation evaluation comprises 10 simulation settings across three task families. Continuous instruction following is measured on the R2R and RxR val-unseen splits~\citep{Krantz2020BeyondTN}. Closed-vocabulary ObjectNav is evaluated on MP3D and HM3D v1/v2, and open-vocabulary generalization is measured on HM3D-OVON. Embodied visual tracking is evaluated on the STT and DT settings of EVT-Bench~\citep{wang2025trackvla}. All navigation benchmarks use the same checkpoint without benchmark-specific fine-tuning.

\begin{figure*}[!t]
    \centering
    \includegraphics[width=\textwidth,pagebox=cropbox]{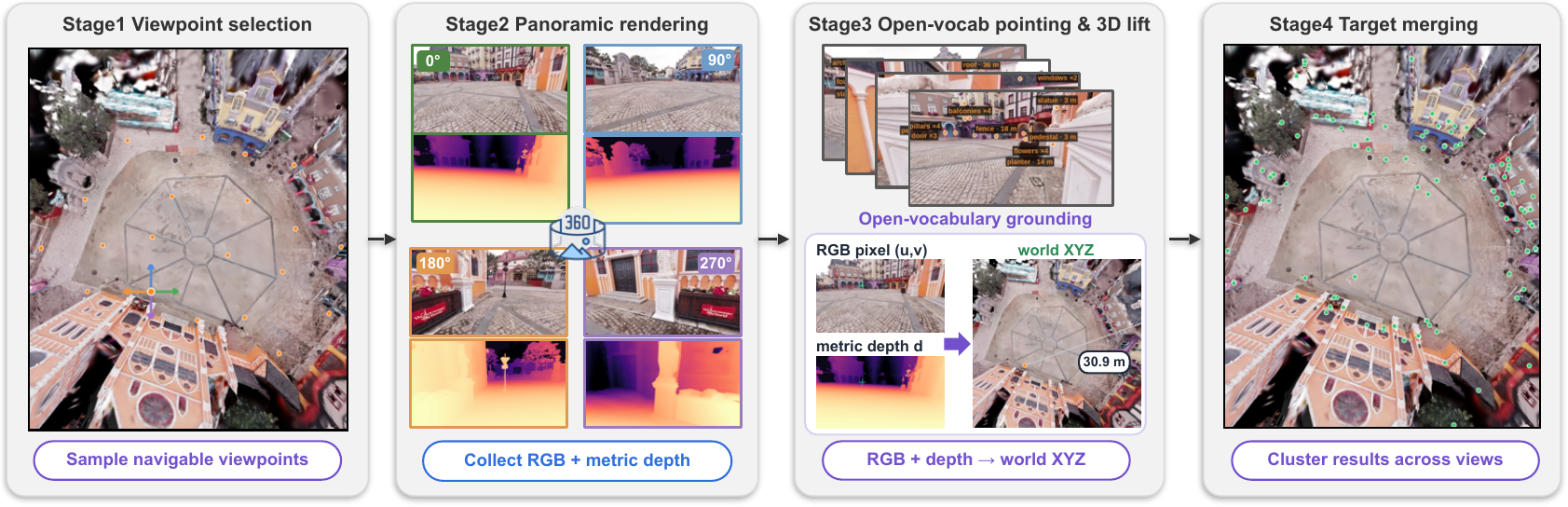}
    \caption{\textbf{Automatic pre-annotation pipeline for INSIGHT-Bench.} The pipeline first samples navigable viewpoints (orange dots) and renders RGB and metric-depth observations in 4 headings. Molmo2 then produces open-set image-space pointing predictions, which are lifted into 3D and merged across views. The green dots denote the resulting spatially consistent 3D target instances.}
    \label{fig:insight-annotation}
\end{figure*}

\begin{figure*}[!t]
    \centering
    \includegraphics[width=\textwidth,pagebox=cropbox]{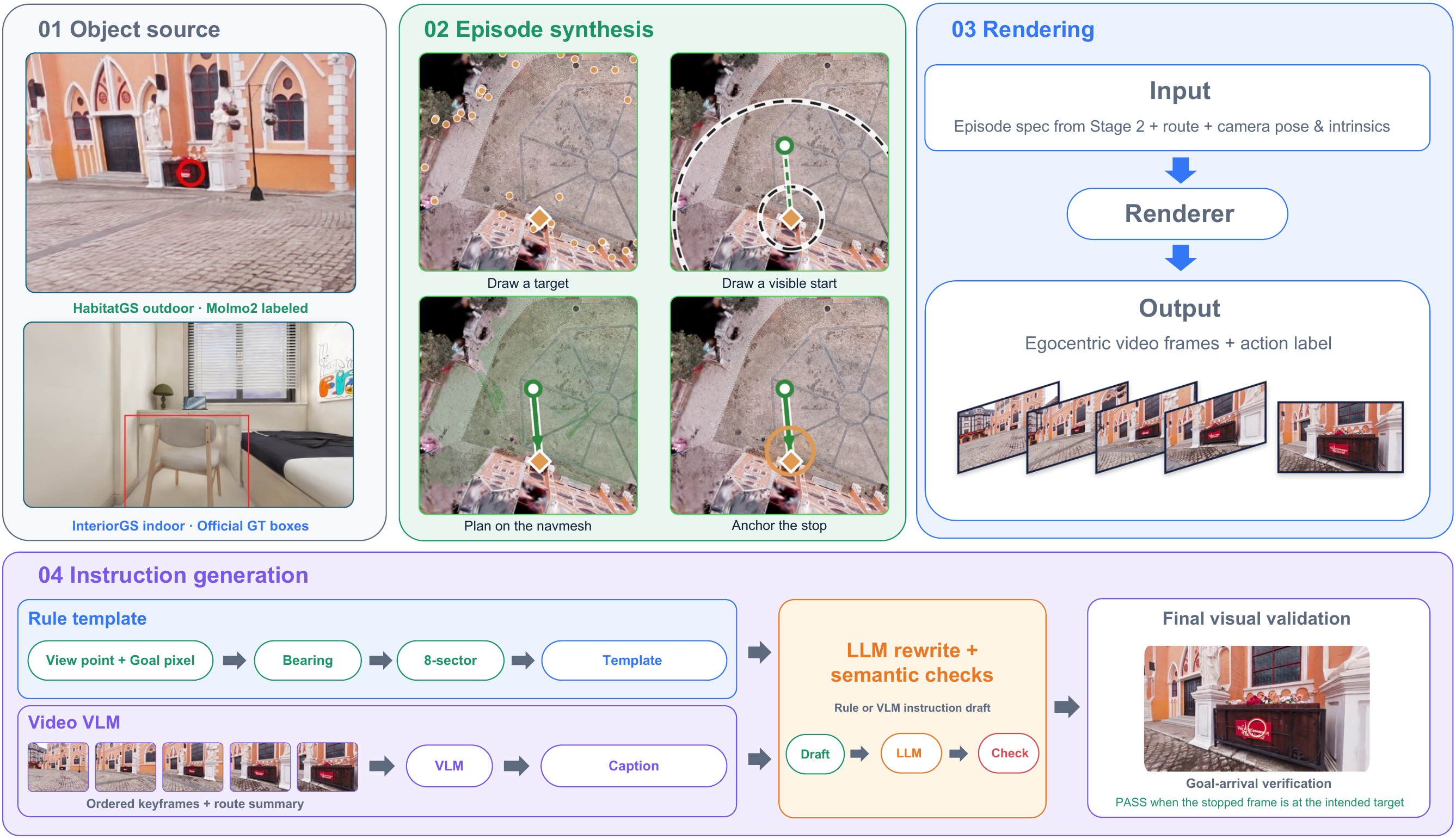}
    \caption{\textbf{Automatic data-generation and instruction-labeling pipeline for INSIGHT-Bench.} All scene sources are converted to a common 3D target inventory; we then draw a target and a start pose from which it is visible in at least one offline inventory view, plan a route on the navigation mesh, anchor the stop, and render an egocentric clip with action labels. The evaluated policy remains monocular, and the target need not appear in its initial forward-facing observation. Instructions are drafted by either a rule-template route or a video-VLM route, rewritten by an LLM under semantic checks, and confirmed by a visual goal-arrival check on the final frames.}
    \label{fig:insight-data-generation}
\end{figure*}

\subsection{INSIGHT-Bench}

\begin{figure*}[!t]
    \centering
    \includegraphics[width=\textwidth,pagebox=cropbox]{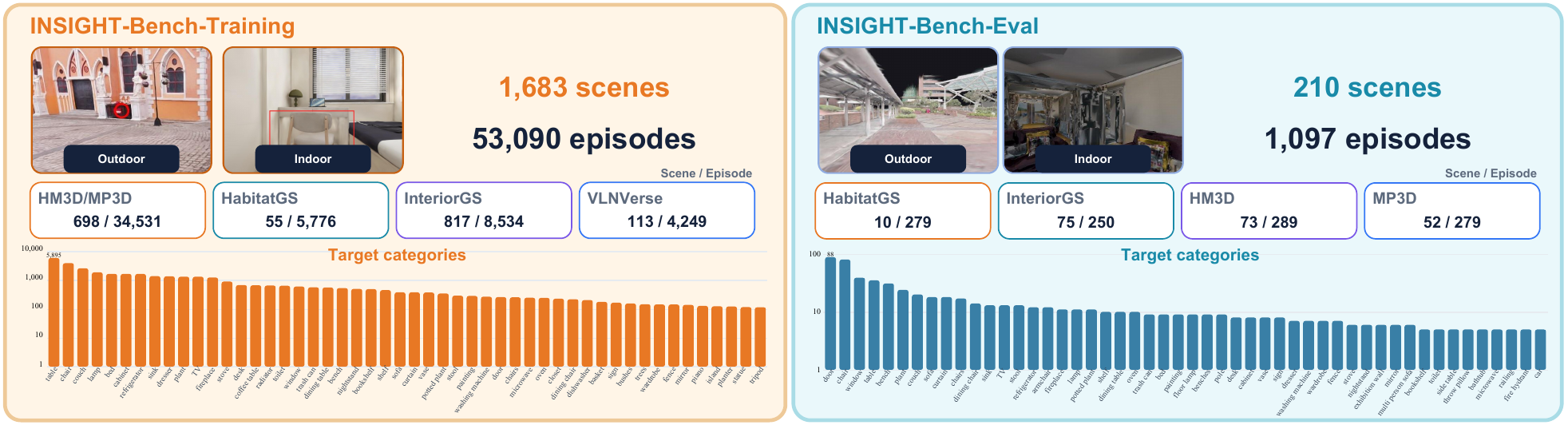}
    \caption{\textbf{Composition of INSIGHT-Bench.} The training split contains 1,683 scenes and 53,090 episodes collected from HM3D/MP3D, Habitat-GS, InteriorGS, and VLNVerse, whereas the evaluation split contains 210 scenes and 1,097 episodes from Habitat-GS, InteriorGS, HM3D, and MP3D. The lower histograms show the Top-50 navigation-target frequencies on a logarithmic scale.}
    \label{fig:insight-bench-overview}
\end{figure*}

INSIGHT-Bench extends the data distribution with heterogeneous mesh-based and 3D Gaussian-splatting environments. It contains 1,683 scenes and 53,090 training episodes, together with 210 scenes and 1,097 evaluation episodes. Fig.~\ref{fig:insight-bench-overview} details the source composition of both splits.

\subsubsection{Pre-annotation}

Scene sources enter the target-inventory stage through two routes. Unlabeled Habitat-GS captures, together with HM3D/MP3D and VLNVerse, use the automatic multi-view pre-annotation pipeline in Fig.~\ref{fig:insight-annotation}, whereas InteriorGS bypasses visual discovery and directly supplies official ground-truth boxes. Both routes are converted to the same 3D target representation before episode synthesis. For these sources, the orange dots in Stage~1 mark sampled navigable viewpoints, each of which provides RGB and metric-depth observations in 4 headings. We use Molmo2~\citep{clark2026molmo2} for open-set pointing, localizing candidate targets in the rendered views from open-vocabulary object prompts. Metric depth lifts every image-space prediction into 3D. Stage~4 groups category-compatible observations with nearby 3D locations and retains a merged instance only when it is supported by at least two observations from at least two distinct viewpoints and its 3D localization spread is below $0.6\,\mathrm{m}$. Each accepted cluster is represented as one target instance, shown by a green dot in the figure. This cross-view consistency gate removes single-view detections and geometrically unstable matches before episode generation. The resulting target geometry is projected back to the agent view to construct object-point supervision, while navigable free space provides the corresponding affordance-point target.

\subsubsection{Data Generation and Instruction Labeling}

\begin{figure*}[!t]
    \centering
    \includegraphics[width=\textwidth,pagebox=cropbox]{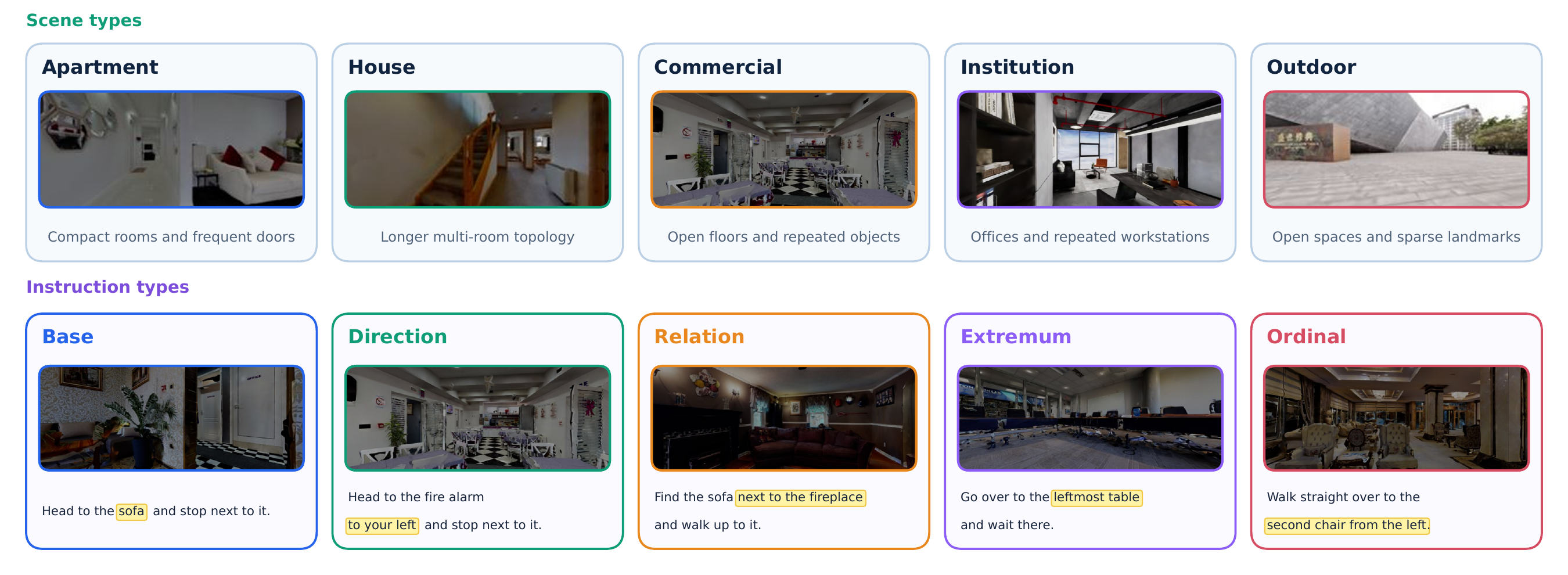}
    \caption{\textbf{Two-axis diagnostic taxonomy of INSIGHT-Bench.} The scene axis groups environments by dominant function and layout, independent of their source dataset. The instruction axis identifies the spatial mechanism required to resolve the goal.}
    \label{fig:insight-bench-taxonomy}
\end{figure*}

We then sample a target together with a starting pose from which the target is visible in at least one offline inventory view. This construction-time constraint enables geometric labeling but does not guarantee target visibility to the policy: evaluation uses only the forward-facing monocular stream, and the target need not appear in the initial observation. We plan a route on the navigation mesh, anchor the stopping location, and render egocentric video frames with action labels, as shown in Fig.~\ref{fig:insight-data-generation}. This common interface makes mesh-based simulators and Gaussian-splatting scenes compatible with the same trajectory format while increasing visual diversity without introducing a scene-specific model component.

Each instruction is then produced by one of two alternative drafting routes, followed by a shared rewriter and a final visual check. The rule-based route reuses evidence that the target inventory already carries: every grounded target records the camera view in which it was detected and the horizontal position of its goal pixel inside that view. Together these determine the direction of the target relative to the agent's starting heading, so the direction word is read off the recorded geometry rather than inferred from appearance.

The wording is then chosen in two steps. The detected view first fixes the turn family: a target found in the left view is reached by turning left, and a target found in the rear view by turning around. The goal pixel then only refines the phrase within that family, separating \emph{ahead} from \emph{front-left} and \emph{front-right}, or \emph{behind-left} from \emph{behind-right}. The resulting sectors are therefore view-aware rather than uniform angular bins, and letting the view dominate is what keeps the wording faithful: a target seen in the rear view whose bearing falls inside the left range is behind the agent on its left, and describing it as a left turn would send the agent the wrong way. Each sector carries a turn-verb phrasing and a locative phrasing, so that ``Turn left and go to the chair and stop.'' and ``The chair is on your left; go to it and stop.'' both occur. Episodes whose planned route departs substantially from the straight line to the target always take the locative form, because it describes the direction of the target, whereas a turn verb is read as describing the path that is actually executed.

The video-conditioned route describes the same episode from its recording instead of its geometry. 6 frames evenly spaced over the rendered clip are submitted together with a summary of the executed trajectory to Seed2.0~\citep{bytedance2026seed2}, which is asked to describe the motion along the route rather than only the static position of the target. Its drafts therefore mention landmarks passed on the way and are phrased more freely than any template allows.

Both routes emit drafts rather than final instructions. The same model then rewrites every draft to broaden its vocabulary without changing what it denotes, and each rewrite is accepted only if it passes the semantic checks illustrated in the image: it must keep the head noun of the target, keep the direction implied by its sector, and never introduce the opposite direction. A rewrite that fails falls back to the corresponding source draft, so no episode is lost to rewriting.

Finally, a visual arrival check closes the loop on the rendered episode. The same pointing model is applied to the last frames of the rendered clip (by default the 1st, 3rd, 6th, and 10th frames counted back from the end) and asked to locate the instructed target; an episode is confirmed as soon as any frame contains it, and episodes in which none does are flagged as suspect for review. The 4 components play complementary roles across the benchmark: the rule route supplies exact egocentric direction, the video route supplies route-level language, the rewriter supplies lexical variety, and the checks preserve target identity and trajectory semantics.

\subsubsection{Diagnostic Taxonomy}

\input{table/insight_bench_taxonomy}

We organize INSIGHT-Bench along two independent axes. Each episode inherits one of 5 scene-level labels from the frozen scene-class mapping: apartments contain compact rooms and frequent doors; houses have longer multi-room topologies; commercial scenes contain open floor plans and repeated object instances; institutions emphasize corridors and repeated rooms or workspaces; and outdoor scenes contain open traversable regions with comparatively sparse landmarks. These labels describe functional layout rather than the source dataset or rendering representation.

Instruction labels are assigned at the episode level from the geometric proof used to instantiate the target, rather than inferred post hoc from surface wording. A \emph{base} instruction names a uniquely resolvable target without a spatial modifier; \emph{direction} adds an egocentric side or bearing; \emph{relation} identifies the target through a unique anchor object; \emph{extremum} selects an argmin or argmax instance such as the nearest or leftmost target; and \emph{ordinal} selects a ranked instance from a stable ordered set. Fig.~\ref{fig:insight-bench-taxonomy} illustrates both axes, and Tab.~\ref{tab:insight-bench-taxonomy} reports their complete $5\times5$ episode distribution.

The taxonomy turns a single aggregate success rate into a diagnostic result. Row-wise differences expose sensitivity to scene function and layout, column-wise differences isolate the underlying language mechanism, and individual cells reveal interactions such as ordinal references in repeated commercial layouts. The lower counts for relation and ordinal episodes reflect stricter uniqueness and ordering gates, rather than silent truncation. We therefore treat aggregate success rate as an overall health indicator and use the scene, instruction, and cross-category results for substantive conclusions.

\subsubsection{Benchmark Statistics}

The training and evaluation scene sets are strictly disjoint. Specifically, none of the 210 evaluation scenes overlaps with any of the 1,683 training scenes across data sources. The Habitat-GS captures are partitioned contiguously into 55 training and 10 evaluation scenes, InteriorGS contributes 817 training and 75 evaluation scenes, and the HM3D partition is fixed by a frozen scene allowlist that every collection run consumes. The training split averages 31.5 episodes per scene, whereas the evaluation split averages 5.2. Because the benchmark combines conventional indoor meshes with Gaussian-splatting reconstructions, it also tests whether a policy trained under a unified visual interface transfers across rendering representations and scene types.

%% file: table/insight_bench_taxonomy.tex
\begin{table*}[t]
    \centering
    \small
    \setlength{\tabcolsep}{5.5pt}
    \caption{\textbf{Episode distribution of the INSIGHT-Bench evaluation split across two diagnostic axes.} Rows group scenes by dominant function and layout; columns group instructions by the spatial mechanism required to identify the target. ``Scenes'' counts unique evaluation environments, while the remaining cells count evaluation episodes.}
    \label{tab:insight-bench-taxonomy}
    \begin{tabular}{lrrrrrrr}
        \toprule
        Scene type & Scenes & Episodes & Base & Direction & Relation & Extremum & Ordinal \\
        \midrule
        Apartment & 120 & 239 & 50 & 50 & 50 & 50 & 39 \\
        House & 61 & 216 & 50 & 50 & 31 & 50 & 35 \\
        Commercial & 10 & 195 & 42 & 40 & 30 & 50 & 33 \\
        Institution & 11 & 219 & 50 & 49 & 32 & 50 & 38 \\
        Outdoor & 8 & 228 & 45 & 50 & 50 & 50 & 33 \\
        \midrule
        Total & 210 & 1,097 & 237 & 239 & 193 & 250 & 178 \\
        \bottomrule
    \end{tabular}
\end{table*}

%% file: section/training_recipe.tex
\section{Training Recipe}

\subsection{Embodied-Reasoning Mid-training}

Although the pretrained VLM already encodes broad semantic and spatial priors, these priors are not consistently exposed in the forms required for embodied decision-making. We therefore begin with embodied-reasoning (ER) mid-training, which specializes the backbone before introducing navigation actions. As summarized in Fig.~\ref{fig:two-stage-vqa-data}, Stage I uses a task-balanced mixture spanning spatial reasoning, general VQA, image pointing, video QA, multi-image reasoning, situated 3D QA, affordance and failure understanding, and scene description. These complementary tasks jointly train the model to localize visual evidence, reason across viewpoints and time, identify feasible free space, and anticipate the consequences of embodied interactions. Importantly, this stage operates entirely through the native autoregressive interface and introduces no navigation-specific prediction head. We refer to the resulting embodied-reasoning checkpoint as LightNav-ER and use it to initialize subsequent navigation alignment.

Specialization alone can narrow the general visual-language competence inherited from the backbone. We therefore adopt a specialize-then-retain curriculum: during the subsequent supervised stage, a rebalanced ER mixture is rehearsed together with navigation data. This second stage preserves the pointing, grounding, and spatial-reasoning skills acquired during ER mid-training while aligning them with embodied action generation. The resulting curriculum establishes a common perceptual and reasoning basis before cross-task navigation supervision is introduced.

For ER mid-training, we use a learning rate of $1\times10^{-5}$ and a warmup ratio of $0.01$. The global batch size is $128$, and the maximum sequence length is $10{,}240$ tokens. This stage is trained on NVIDIA H100 GPUs and consumes approximately $170$ H100 GPU-hours of compute.

\subsection{Supervised Fine-tuning}

We next perform supervised fine-tuning (SFT) to align the LightNav-ER checkpoint with the unified navigation token space. The task-balanced optimization mixture in Fig.~\ref{fig:sft-data-composition} combines retained ER and VQA data with instruction following, object navigation, and embodied visual tracking. Navigation examples are serialized using the same target structure across tasks: one $\langle\mathrm{apos}_{i}\rangle$ token and one $\langle\mathrm{opos}_{i}\rangle$ token provide the latent spatial trace, followed by three RVQ tokens specifying the short-horizon trajectory. Auxiliary reasoning examples and navigation trajectories are optimized with the shared causal language-model objective described above, allowing the shared backbone and output head to learn across tasks, scenes, and embodiments. The navigation mixture additionally includes DAgger-collected examples~\citep{ross2011dagger}, exposing the policy to states induced by its own predictions and reducing the train-deployment state-distribution gap.

We train with a learning rate of $1.5\times10^{-5}$ and a warmup ratio of $0.01$. The global batch size is $320$, and the maximum sequence length is $8{,}192$ tokens. Training is conducted on NVIDIA H100 GPUs and consumes approximately $950$ H100 GPU-hours of compute. Together with dynamic sequence packing, this configuration accommodates long visual histories while retaining a large and diverse global batch.

\subsection{Online RL Post-training}

\begin{figure*}[!t]
    \centering
    \includegraphics[width=\textwidth,pagebox=cropbox]{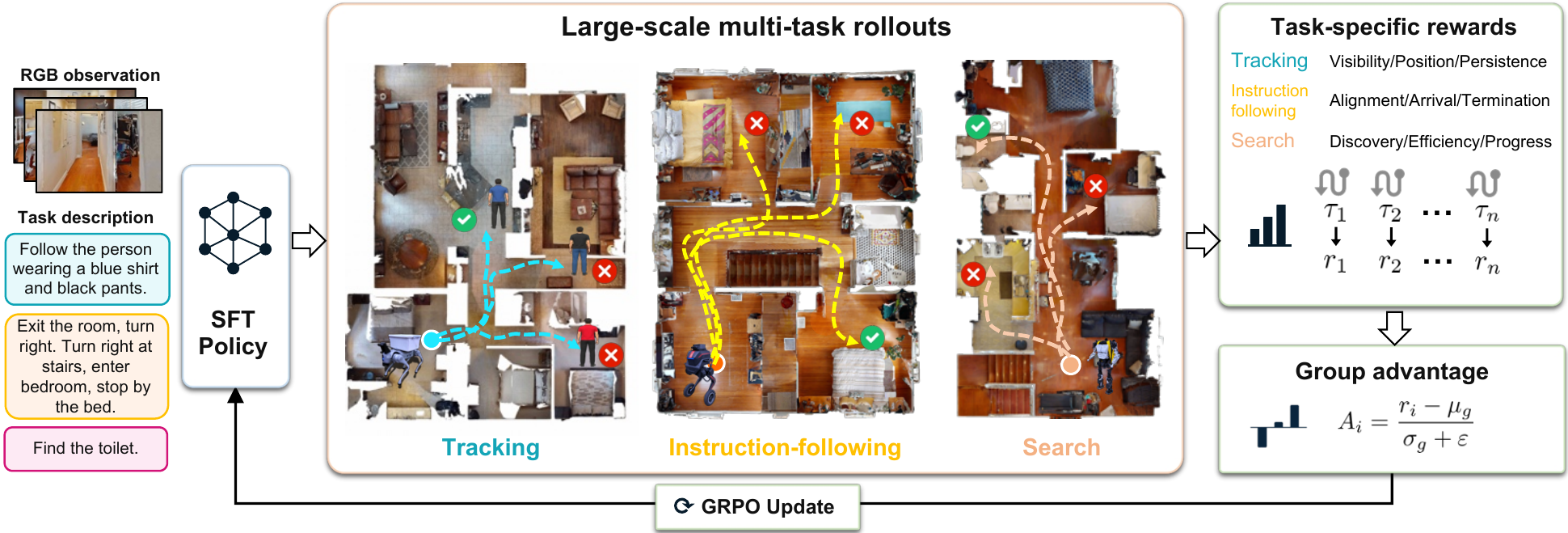}
    \caption{\textbf{Online multi-task reinforcement-learning pipeline.} We initialize online optimization from the supervised policy and collect large-scale rollouts for all navigation tasks. Task-specific rewards evaluate visibility, position, and persistence for EVT; alignment, arrival, and termination for instruction following; and discovery, efficiency, and progress for object goal navigation. Rewards are normalized within each rollout group to compute advantages for GRPO policy updates.}
    \label{fig:rl-pipeline}
\end{figure*}

Although DAgger augments SFT with policy-induced states, the resulting objective
remains token-level imitation and does not directly optimize the closed-loop
behaviors that determine task success. We therefore introduce online
reinforcement learning to optimize complete policy rollouts against task-level
objectives, as shown in Fig.~\ref{fig:rl-pipeline}. This stage refines long-horizon
behavior, including sustained tracking, route-consistent goal reaching,
efficient search, and appropriate termination. The same backbone, token
interface, and rollout machinery support embodied visual tracking,
instruction-following VLN, and open-vocabulary object goal navigation; only the
task-specific reward function differs.

\subsubsection{Problem Setup and Group-Relative Objective}

An episode is a trajectory
$\tau=(\mathbf{s}_1,\mathbf{a}_1,\ldots,\mathbf{s}_T,\mathbf{a}_T)$. Its state
$\mathbf{s}_t$ pairs the instruction $\mathcal{I}$ with the compressed history
$\mathcal{O}_{1:t}$, and its action $\mathbf{a}_t$ is the token block emitted at
step $t$: the dual-channel pointing prefix followed by three RVQ action
tokens. 

We optimize with Group Relative Policy Optimization
(GRPO)~\citep{shao2024deepseekmath}, which replaces a learned value function by
a group baseline. For each episode seed we roll out $G$ independent trajectories,
score each with a single scalar $R(\tau^{(g)})$, and take the within-group
standardized reward as the advantage,
\begin{equation}
    \begin{aligned}
        A^{(g)} &= \frac{R(\tau^{(g)})-\mu_R}{\sigma_R+\epsilon_{\mathrm{num}}} \\[0.35em]
        \mu_R &= \frac{1}{G}\sum_{g}R(\tau^{(g)})
    \end{aligned}
    \label{eq:grpo-advantage}
\end{equation}
where $\sigma_R$ is the group standard deviation and $\epsilon_{\mathrm{num}}=10^{-6}$ is a numerical stability constant. Standardizing inside the group
removes the scene-difficulty offset, so only the ranking of the $G$ attempts on
the same start state carries gradient.

Each task is scored by one terminal scalar that is assigned to every decision
step, $r_t=R(\tau)$ for all $t$, so all tokens of a trajectory share the
advantage $A^{(g)}$. Per-step shaping would instead need a hand-designed
potential defined three times over three incompatible geometries. The surrogate is applied at token
level, with $\rho$ the importance ratio of a response token,
\begin{equation}
\begin{split}
    \mathcal{L}_{\mathrm{RL}}=-\,\mathbb{E}\Big[&\min\big(\rho\,A,\ \operatorname{clip}(\rho,1-\varepsilon,1+\varepsilon)\,A\big)\Big]\\
    &+\beta\,D_{\mathrm{KL}}\!\left(\pi_\theta\,\Vert\,\pi_{\mathrm{ref}}\right),
\end{split}
    \label{eq:rl-loss}
\end{equation}
where $\pi_{\mathrm{ref}}$ is the frozen supervised policy and $\beta$ anchors
the update to it.

\subsubsection{Rollout Infrastructure and Training-Set Construction}

Online RL is bounded by simulation throughput rather than by gradient
computation, so we decouple simulation from optimization. Each simulator is an independent process
holding one resident scene, and action generation is served by a separate
inference server. A scene hash assigns all \(G\) rollouts from the same episode seed to a simulator holding the corresponding scene, thereby avoiding redundant scene loading.

Uniform sampling of the training set would waste most of this budget, since an
episode the supervised policy always solves and one it never solves both yield
$\sigma_R\approx0$ and no gradient. We therefore run $K$ rollouts of every
candidate with the supervised checkpoint and bin it as always-solved, mixed, or
never-solved. Mixed episodes sit on the decision boundary and are the only ones guaranteed to
produce within-group variance, so they dominate the pool. Robostral
Navigate~\citep{bounhar2026robostral} applies a related filter, though it retains
only the tasks its supervised policy fails to solve. We instead keep a small
fraction of both extremes: always-solved episodes preserve acquired behavior,
whereas never-solved episodes provide challenging cases for exploration. The continuous terminal signals described below still rank these failures and thereby support graded trial-and-error learning.

\subsubsection{Task-Specific Terminal Rewards}
\label{sec:rl-rewards}

We define a separate terminal reward for each navigation task, tailored to its objective and success criteria. All distances are reported in meters and bearings in radians.

\paragraph{Embodied visual tracking}
The target here is a moving agent rather than a fixed location, so there is no
goal coordinate to arrive at and no terminating stop for the policy to emit. An
all-zero plan holds the follower in place, and following resumes as soon as the
target moves again. Termination is decided by the environment: the target
finishes its route, the follower loses it beyond recovery, or the step budget is
exhausted. The environment declares success only if the follower is still $1$ to
$3\,\mathrm{m}$ behind the target and oriented towards it at that moment, so
closing in too tightly does not count. What matters over the episode is therefore
sustained visibility at a correct standoff rather than a terminal event. At each step we read the target's visibility $v_t\in\{0,1\}$, bearing
$\alpha_t$, and range $\delta_t$. These give a per-step quality
$q_t = v_t\,c(\alpha_t)\,b(\delta_t)$, whose two position factors are a
flat-topped bearing term and a two-sided range band,
\begin{equation}
    c(\alpha)=
    \begin{cases}
        1, & |\alpha|\le\alpha_0,\\[2pt]
        \exp\!\big(-\tfrac{(|\alpha|-\alpha_0)^2}{2\sigma_\alpha^{2}}\big), & |\alpha|>\alpha_0,
    \end{cases}
    \label{eq:evt-bearing}
\end{equation}
\begin{equation}
    b(\delta)=
    \begin{cases}
        \exp\!\big(-\tfrac{(\delta_{\mathrm{low}}-\delta)^2}{2\sigma_{\mathrm{near}}^{2}}\big), & \delta<\delta_{\mathrm{low}},\\[2pt]
        1, & \delta_{\mathrm{low}}\le\delta\le\delta_{\mathrm{high}},\\[2pt]
        \exp\!\big(-\tfrac{(\delta-\delta_{\mathrm{high}})^2}{2\sigma_{\mathrm{far}}^{2}}\big), & \delta>\delta_{\mathrm{high}}.
    \end{cases}
    \label{eq:evt-band}
\end{equation}
The bearing factor uses a dead zone $\alpha_0=0.14\,\mathrm{rad}$ and a width
$\sigma_\alpha=0.35\,\mathrm{rad}$. The range band is
$[\delta_{\mathrm{low}},\delta_{\mathrm{high}}]=[1.5,\,3.0]\,\mathrm{m}$, with
$\sigma_{\mathrm{near}}=0.35\,\mathrm{m}$ and
$\sigma_{\mathrm{far}}=1.0\,\mathrm{m}$. Inside the flat top and the flat band
the factors are exactly $1$, so fine-grained jitter produces no group variance.
A second per-step term scores the motion rather than the view. Let
$m_t\in[0,1]$ measure how closely the 10-waypoint plan emitted at step $t$
matches a privileged oracle plan from the same state. Writing
$\mathds{1}[\cdot]$ for the indicator function, combining the two per-step terms
with a collision charge gives
\begin{equation}
\begin{aligned}
    R_{\mathrm{EVT}} =\;& \mathds{1}[\text{success}]
    - 0.5\times \mathds{1}[\text{collision}]\\
    &+ \frac{1}{T_{\mathrm{norm}}}\sum_{t=1}^{T}\big(0.7\,q_t + 0.3\,m_t\big).
\end{aligned}
    \label{eq:reward-evt}
\end{equation}

Persistence comes from the normalizer
\begin{equation}
    T_{\mathrm{norm}}=
    \begin{cases}
        T, & \text{natural end},\\[2pt]
        \max(T,\,T_0), & \text{early end},
    \end{cases}
    \label{eq:evt-norm}
\end{equation}
a natural end being the target finishing or the budget being reached, an early
end the target being lost or a collision. Averaging over executed steps alone would reward early termination, since a
follower that crashes early stops the clock while its average is high. Charging an early end against the constant
horizon $T_0=300$ steps counts the unexecuted steps as zero quality, which turns
average quality into persistence.

\paragraph{Instruction following}
Let $d_T$ be the geodesic distance from the final pose to the goal, and
$\mathrm{nDTW}$ the normalized dynamic-time-warping similarity between the
executed and the annotated reference path. With a success radius of $3.0\,$m we
use
\begin{equation}
\begin{aligned}
    R_{\mathrm{VLN}} =\;& \big(1+\mathrm{nDTW}\big)\,\mathds{1}[\text{success}]\\
    &+\exp\!\Big(-\tfrac{\max(d_T,\,d_{\mathrm{clip}})^{2}}{\sigma_d^{2}}\Big)
    -0.25\times \mathds{1}[\text{timeout}].
\end{aligned}
\label{eq:reward-vln}
\end{equation}
The first term pays arrival once, and pays up to twice as much when the executed
path also kept alignment with the described route. That bonus prevents a
shortcut that reaches the goal while ignoring the instruction. The Gaussian kernel, of width $\sigma_d=3.0\,\mathrm{m}$, grades that arrival on
both sides of the success boundary and ranks a near miss above a distant stop.
The clip $d_{\mathrm{clip}}=2.0\,\mathrm{m}$ flattens it inside that radius.
A timeout is an episode that exhausts the budget of $T_0=300$ decision steps
without ever calling a stop.

\paragraph{Object goal navigation}
Object goal navigation names a category instead of a route, so path fidelity is
meaningless and only reaching the object matters. Let $d_0$ be the initial
geodesic distance to the goal and $\tilde{\ell}$ the length of the executed path.
We write $\mathrm{PL}=d_0/\max(d_0,\,\tilde{\ell})$ for the per-episode
path-length efficiency, the quantity that SPL averages over a
dataset~\citep{anderson2018evaluation}. With a success radius of $1.0\,$m we use
\begin{equation}
\begin{aligned}
    R_{\mathrm{OBJ}} =\;& \big(1+0.25\,\mathrm{PL}\big)\,\mathds{1}[\text{success}]\\
    &+0.25\,\exp\!\Big(-\tfrac{\max(d_T,\,d_{\mathrm{clip}})^{2}}{\sigma_d^{2}}\Big).
\end{aligned}
\label{eq:reward-obj}
\end{equation}
The first term mirrors Eq.~\eqref{eq:reward-vln}. Arrival is paid once, and paid
more when the object was reached along a short path, which discourages
exhaustive sweeping of the building. Only the measured quality differs: path
efficiency here, path fidelity there, because object navigation prescribes no
route. The graded proximity term then carries the whole
failure population, ranking a run that halved its distance to the object above
one that never left the starting room. Without it every failure would share one
reward value and contribute nothing to Eq.~\eqref{eq:grpo-advantage}. Because that radius is $1.0\,$m rather than the $3.0\,$m used for
instruction following, the clip tightens to $d_{\mathrm{clip}}=1.0\,\mathrm{m}$,
while $\sigma_d$ is unchanged.

\subsubsection{Optimization Details}

Each iteration draws $B$ episode seeds with pairwise-distinct scenes and rolls
out $G$ trajectories per seed. Because each decision step constitutes one
training sample, a single iteration generates far more samples than one update
needs. We therefore cap the number retained per episode and select them by event
stratification. The first and last decisions are always kept, together with
decisions carrying a discrete event such as stop emission, stuck detection,
collision, or a large-angle turn, and their immediate neighbors. Any remaining
quota is spread uniformly over the timeline. Uniform subsampling would discard
these decisions in proportion to their rarity, yet they are where the terminal
reward is actually earned.

The group size is $G=8$ and each iteration draws $B=32$ seeds, giving $256$
episodes per update. The clip range is $\varepsilon=0.2$, the KL coefficient is
$\beta=0.01$, and the learning rate is $1\times10^{-6}$. Training runs on a
single node with 8 H100 GPUs.

%% file: section/experiments.tex
\section{Experiments}

\subsection{Experimental Setup}

\subsubsection{Benchmarks}

We evaluate spatial intelligence, simulated navigation, and real-world transfer. For spatial intelligence, LightNav-ER is tested on 8 embodied-reasoning benchmarks: Point-Bench~\citep{cheng2025pointarena}, RefSpatial~\citep{zhou2025roborefer}, the POI and VQA tracks of RoboSpatial~\citep{song2024robospatial}, Where2Place~\citep{yuan2024robopoint}, CV-Bench~\citep{tong2024cambrian}, ERQA~\citep{geminirobotics2025}, and EmbSpatial~\citep{du2024embspatial}. For navigation, a single shared \ours checkpoint is evaluated without benchmark-specific fine-tuning on 10 public simulation settings spanning three task families and INSIGHT-Bench. Instruction following is measured on the R2R~\citep{anderson2018vision} and RxR~\citep{ku2020room} val-unseen splits in VLN-CE~\citep{Krantz2020BeyondTN}; object goal navigation is evaluated on MP3D~\citep{chang2017matterport3d}, HM3D v1/v2~\citep{ramakrishnan2021hm3d}, and HM3D-OVON~\citep{yokoyama2024hm3d}; and embodied visual tracking is evaluated on the STT and DT splits of EVT-Bench~\citep{wang2025trackvla}. We further deploy the same checkpoint on four physical robot embodiments to assess zero-shot real-world deployment beyond simulation.

INSIGHT-Bench is evaluated under a shared deployment protocol. All models see the same 1,097 episodes from 210 scenes, receive a $120^{\circ}$, $480{\times}270$ forward RGB stream from a camera at $1.0\,\mathrm{m}$ height, and are given a budget of 300 actions. An episode succeeds only if the model stops within the goal radius and the target lies inside the $120^{\circ}$ field of view of the final frame. The radius is $2.0\,\mathrm{m}$ for indoor scenes and $3.0\,\mathrm{m}$ for outdoor scenes, because outdoor environments are considerably larger and their navigable targets are correspondingly farther apart.

\subsubsection{Baselines}

For embodied reasoning, we compare with two general-purpose 4B VLMs, Qwen3-VL and the post-trained Qwen3.5-4B checkpoint~\citep{qwen2026qwen35}, and the spatially specialized 8B Molmo2-ER model. Navigation comparisons cover three representative families: modular systems that separate perception, mapping, and planning; task-specific end-to-end policies, including methods with learned waypoint predictors; and generalist VLM/VLA navigation policies. For INSIGHT-Bench, we additionally run 6 open-source navigation policies through a common action interface: JanusVLN~\citep{zeng2025janusvln}, NaVid~\citep{zhang2024navid}, Uni-NaVid~\citep{zhang2024uni}, TAMP-Nav~\citep{feng2026tampnav}, InternVLA-N1~\citep{internvla2025}, and StreamVLN~\citep{wei2025streamvln}. Each policy receives the shared forward stream through an adapter that may center-crop it to the model's native training field of view, as used by NaVid and StreamVLN. TAMP-Nav was designed for 4-camera $360^{\circ}$ observations and depth-assisted pixel-to-3D execution, but we restrict its visual input to the shared forward view. InternVLA-N1 follows its released RGB-only path with constant depth. The episode set, simulator, action budget, and success criterion are identical across these runs. Every navigation result explicitly identifies the use of single-view RGB, panoramic or multi-camera RGB, depth, and odometry.

\subsubsection{Metrics}

For embodied reasoning, we report each benchmark's primary score as a percentage and compute an unweighted macro average across all 8 benchmarks only when every result is available. For continuous VLN, we report navigation error (NE), oracle success rate (OS), success rate (SR), success weighted by path length (SPL), and normalized dynamic time warping (nDTW). ObjectNav and HM3D-OVON are evaluated with SR and SPL. INSIGHT-Bench is evaluated with SR, SPL, and terminal NE. EVT-Bench reports success rate (SR), tracking rate (TR) and collision rate (CR). Higher values indicate better performance for all metrics except NE and CR, as marked by the arrows in each table.

\subsection{Embodied Reasoning Benchmark Evaluation}

\input{table/er_bench}

We first evaluate whether embodied-reasoning (ER) mid-training strengthens the spatial capabilities of the VLM before downstream navigation alignment. We test the 4B ER checkpoint on 8 benchmarks covering language-guided pointing, spatial referring, robotic spatial reasoning, affordance prediction, visual perception, and embodied question answering. Tab.~\ref{tab:er-bench} reports each benchmark's primary score as a percentage, with higher values indicating better performance. The macro average is computed only for models with results on all 8 benchmarks.

LightNav-ER ranks first on 4 of the 8 benchmarks and second on the remaining 4, attaining the highest complete-set average of 67.4. This exceeds the strongest baseline average, achieved by Qwen3-VL-4B, by +4.3 (6.8\%) and the 8B Molmo2-ER average by +4.6 (7.3\%), despite using only 50\% of the parameters. LightNav-ER also outperforms Qwen3.5-4B on all 8 benchmarks. Molmo2-ER remains stronger on Point-Bench, RoboSpatial-VQA, and ERQA, while Qwen3-VL leads on RoboSpatial-POI. The comparison therefore demonstrates broad and balanced spatial competence rather than uniform dominance on every individual task.

Relative to the Qwen3-VL-4B initialization, ER mid-training improves 7 of the 8 benchmark scores and raises the macro average from 63.1 to 67.4, an absolute improvement of +4.3 (6.8\%). The largest gains occur on Where2Place (+12.6; 19.7\%) and RefSpatial (+11.9; 26.2\%), which directly exercise free-space grounding and multi-step spatial referring.

\subsection{Simulation Benchmark Evaluation}

We compare \ours with reported state-of-the-art systems across instruction following, object goal navigation, and embodied visual tracking. To make the sensing assumptions explicit, all comparisons report whether each method uses single-view RGB, panoramic or multi-camera RGB, depth, and odometry inputs.

\subsubsection{Vision-Language Navigation}
\input{table/vln_ce}

As shown in Tab.~\ref{tab:vln-ce}, \ours achieves the strongest monocular R2R result on all 4 metrics. Relative to the best prior monocular entries, SR increases from 66.9 to 68.5 (+1.6; 2.4\%) and SPL from 62.3 to 62.8 (+0.5; 0.8\%), while NE decreases from 4.05 to 3.91 (-0.14~m; 3.5\% reduction). OS reaches 73.7, marginally above Qwen-RobotNav-4B at 73.6. The simultaneous gains in SR, SPL, and NE indicate that the improved goal-reaching reliability is retained under path-efficiency and terminal-precision criteria.

On the longer RxR benchmark, \ours likewise obtains the best monocular NE, SR, and SPL. It improves SR over Qwen-RobotNav-8B from 73.4 to 73.6 (+0.2; 0.3\%) and SPL from 63.5 to 64.5 (+1.0; 1.6\%), while reducing NE from 4.09 to 3.66 (-0.43~m; 10.5\% reduction). Its nDTW of 67.4 remains below DualVLN's 70.0, revealing that the higher success and terminal accuracy do not translate uniformly to trajectory fidelity. Panoramic methods retain small advantages on both datasets, but \ours remains competitive while using only a forward RGB view and no depth or odometry.

\subsubsection{Object Goal Navigation}
\input{table/objectnav}

\ours attains the strongest monocular SR and SPL across all three closed-vocabulary ObjectNav settings in Tab.~\ref{tab:objectnav}, despite using neither depth nor odometry. On MP3D, it raises SR over CogNav from 46.6 to 53.3 (+6.7; 14.4\%) and SPL over VLFM from 17.5 to 21.2 (+3.7; 21.1\%). On HM3D v1, SR increases from 73.7 to 74.5 (+0.8; 1.1\%), while SPL rises from 37.3 to 43.9 (+6.6; 17.7\%). On HM3D v2, \ours improves SR from 77.0 to 77.2 (+0.2; 0.3\%) and SPL from 41.3 to 41.5 (+0.2; 0.5\%).

The same RGB-only policy also exceeds the listed multi-view systems. On MP3D, its SR is 1.1 points above Qwen-RobotNav-4B and its SPL is 3.5 points above Qwen-RobotNav-8B. On HM3D v1, it outperforms the depth- and odometry-assisted WMNav by +16.4 SR and +12.7 SPL; on HM3D v2, it exceeds the best multi-view entries by +1.6 SR and +8.5 SPL. These comparisons rule out a wider field of view or privileged geometry as the source of the advantage.

\input{table/hm3d_ovon}

Open-vocabulary evaluation exhibits the same pattern (Tab.~\ref{tab:hm3d-ovon}). On val seen categories, \ours improves the strongest prior monocular SR from 55.0 to 55.3 (+0.3; 0.5\%). The SR gain widens to +9.6 (21.3\%) on val synonyms and +6.2 (15.2\%) on val unseen categories. SPL also rises from 23.6 to 31.2 (+7.6; 32.2\%) on seen categories and from 21.8 to 29.6 (+7.8; 35.8\%) on synonyms. On unseen categories, SPL increases from 19.8 to 24.2 (+4.4; 22.2\%). These margins are measured against monocular methods that may additionally use depth and odometry, making the result particularly notable for a single-RGB-input policy.

\subsubsection{INSIGHT-Bench}

\input{table/insight_bench_overall}

As shown in Tab.~\ref{tab:insight-bench-overall}, \ours attains the best result on every aggregate metric. It improves SR over JanusVLN from 27.4 to 43.7 (+16.3; 59.5\%) and SPL from 24.0 to 41.5 (+17.5; 72.9\%), and reduces NE from NaVid's 4.25~m to 3.88~m (-0.37~m; 8.7\%).

\input{table/insight_bench_breakdown}

The instruction axis of Tab.~\ref{tab:insight-bench-breakdown} shows where that advantage is concentrated. \ours leads every instruction type, and the margin is largest on Direction, which rises from NaVid's 29.7 to 57.7 (+28.0; 94.3\%). Direction is also the only type on which \ours exceeds its own Base score of 45.1, whereas all 6 baselines fall below their Base score once an egocentric bearing is added. This is consistent with the route-conditioned supervision of Sec.~\ref{sec:data-and-benchmarks}, in which egocentric bearings are stated explicitly in the rule templates and preserved through rewriting. Extremum remains the hardest type at 37.2, ahead of NaVid by +14.8 (66.1\%), because selecting an argmin or argmax instance requires detecting several candidates before their positions can be compared.

\input{table/evt_bench}

The scene axis follows the same pattern. \ours is strongest in Apartment scenes at 61.1, exceeding the best open-source result by +21.4 (53.9\%), and gains its largest relative margin outdoors, improving on JanusVLN from 16.7 to 34.2 (+17.5; 104.8\%). The outdoor result is consistent with the inclusion of outdoor 3DGS trajectories during training, whereas the evaluated open-source policies are predominantly trained on indoor Habitat environments. Institution is the weakest scene type at 29.2, although it still exceeds the best open-source result by +10.9 (59.6\%); its large spaces and repeated doors, chairs, and workstations make distant recognition and instance disambiguation difficult.

The joint matrix in Fig.~\ref{fig:insight-bench-matrix} shows that the two axes interact rather than contributing independent, uniform penalties. House--Direction reaches 74.0\% (37/50), while Institution--Extremum falls to 18.0\% (9/50), a 56.0-point range against marginal spans of only 31.9 points across scene types and 20.5 points across instruction types. The larger cross-cell span indicates that difficult language compounds scene-specific ambiguity: familiar residential topology and explicit egocentric cues favor House--Direction, whereas repeated instances in large institutional spaces make the global comparison required by Extremum particularly brittle.

\subsubsection{Embodied Visual Tracking}

On EVT-Bench (Tab.~\ref{tab:evt-bench}), \ours achieves the highest SR on both tracking regimes. For single-target tracking, it raises SR from ReferTrack's 89.4 to 91.7 (+2.3; 2.6\%). Under distracted tracking, where the agent must preserve target identity among distractors, the margin increases from 73.3 to 82.6 (+9.3; 12.7\%). \ours also reduces the best prior monocular distracted-tracking CR from 5.51 to 4.62 (-0.89; 16.2\% reduction). ReferTrack retains the highest TR in both regimes and the lowest single-target CR, indicating that persistent visual lock remains a complementary strength of the specialist tracker.

The cross-modality comparison is especially strong under distraction: \ours achieves 82.6 SR, exceeding the best multi-view result, CoMaTrack's 74.2, by +8.4 (11.3\%), while its single-target SR is within 0.4 of CoMaTrack's 92.1. CoMaTrack still attains a lower CR, and ReferTrack retains a higher distracted-tracking TR, so the result establishes stronger episode-level success rather than uniform dominance on every tracking metric. Together with the VLN and ObjectNav results, this supports the use of one compact RGB-only policy across instruction following, search, and following tasks under markedly different sensing and temporal demands.

\subsection{Cross-Domain and Real-World Generalization}

We qualitatively evaluate whether the unified spatial interface transfers beyond the simulation domains used for training. These demonstrations use the same \ours checkpoint without domain-specific fine-tuning. Because the virtual and physical environments do not share a standardized action space or success protocol, we use the rollouts to assess the breadth of transfer rather than to make a quantitative benchmark comparison.

\subsubsection{Cross-Domain Generalization}

\begin{figure*}[!t]
    \centering
    \includegraphics[width=\textwidth,pagebox=cropbox]{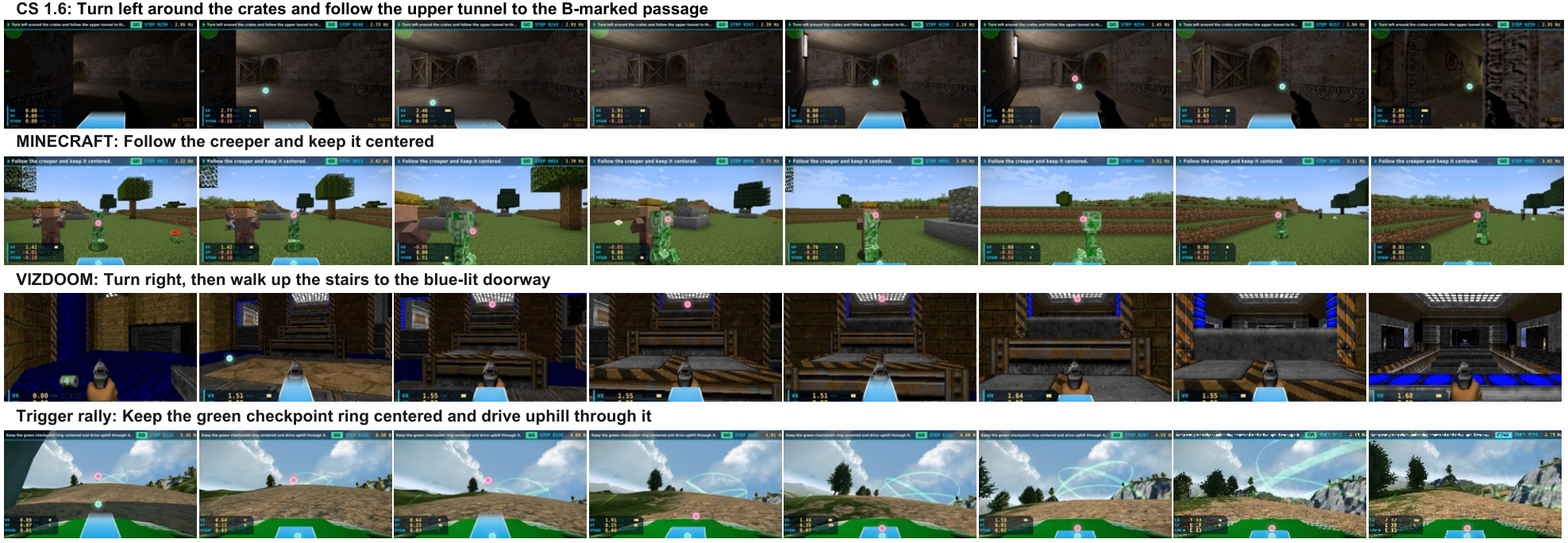}
    \caption{\textbf{Zero-shot generalization across game domains.} The same \ours checkpoint follows language instructions in Counter-Strike 1.6 and VizDoom, tracks a moving target in Minecraft, and performs checkpoint-conditioned driving in Trigger Rally. Cyan and magenta markers visualize the predicted affordance and object points, respectively.}
    \label{fig:game-demo}
\end{figure*}

As shown in Fig.~\ref{fig:game-demo}, \ours operates across 4 game domains with markedly different visual styles, scene structures, and control dynamics. The model follows multi-step spatial instructions in Counter-Strike 1.6 and VizDoom, maintains a moving Creeper as the target in Minecraft, and follows a sequence of checkpoints while driving in Trigger Rally. Across these settings, the affordance point provides a transferable intermediate target in navigable space, while the object point identifies the referred target or landmark when applicable. The coherent rollouts across first-person navigation, target following, and vehicle control indicate that the learned pointing-and-trajectory interface is not tied to the appearance statistics or locomotion dynamics of the training simulators.

\subsubsection{Real-World Generalization}

\begin{figure*}[!t]
    \centering
    \includegraphics[width=1\textwidth,pagebox=cropbox]{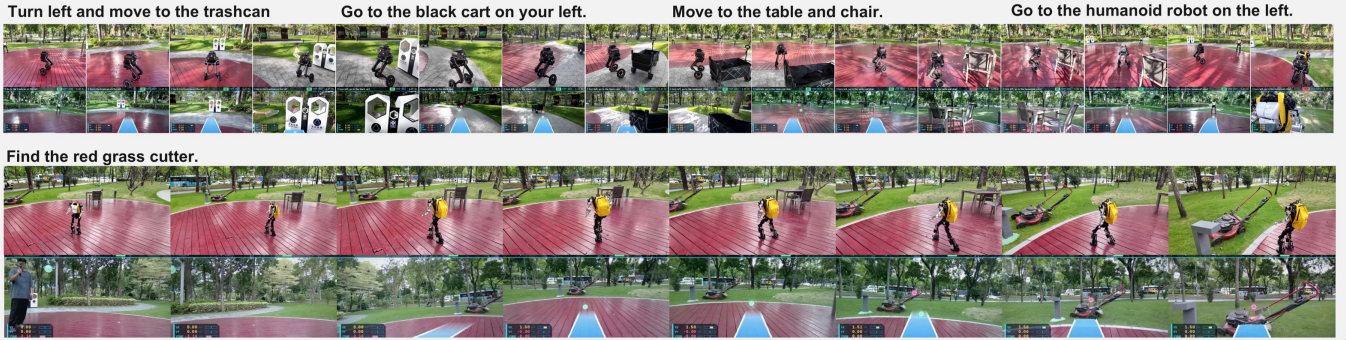}
    \caption{\textbf{Zero-shot real-world generalization across tasks and scenes.} \ours follows people and robotic targets, executes indoor and outdoor navigation instructions, and searches for open-vocabulary objects without model adaptation. Each rollout combines external views of the robot with the corresponding egocentric observations used by the policy.}
    \label{fig:real-world-demo}
\end{figure*}

The same checkpoint is deployed in physical environments without task- or scene-specific adaptation. As shown in Fig.~\ref{fig:real-world-demo}, it supports visual tracking, instruction following, and object search under substantial visual variation. The tracking transfer provides a particularly stringent test of zero-shot generalization. Although tracking supervision contains only human targets, \ours follows previously unseen classes of dynamic targets, including humanoid robots, wheeled robots, and carts, without additional training. Across these rollouts, the model preserves target identity despite changes in viewpoint, background clutter, and illumination, while grounding free-space destinations for navigation and search. These results demonstrate that the learned spatial interface generalizes beyond the scenes, task semantics, and target categories represented during training.

\begin{figure}[!t]
    \centering
    \includegraphics[width=\columnwidth,pagebox=cropbox]{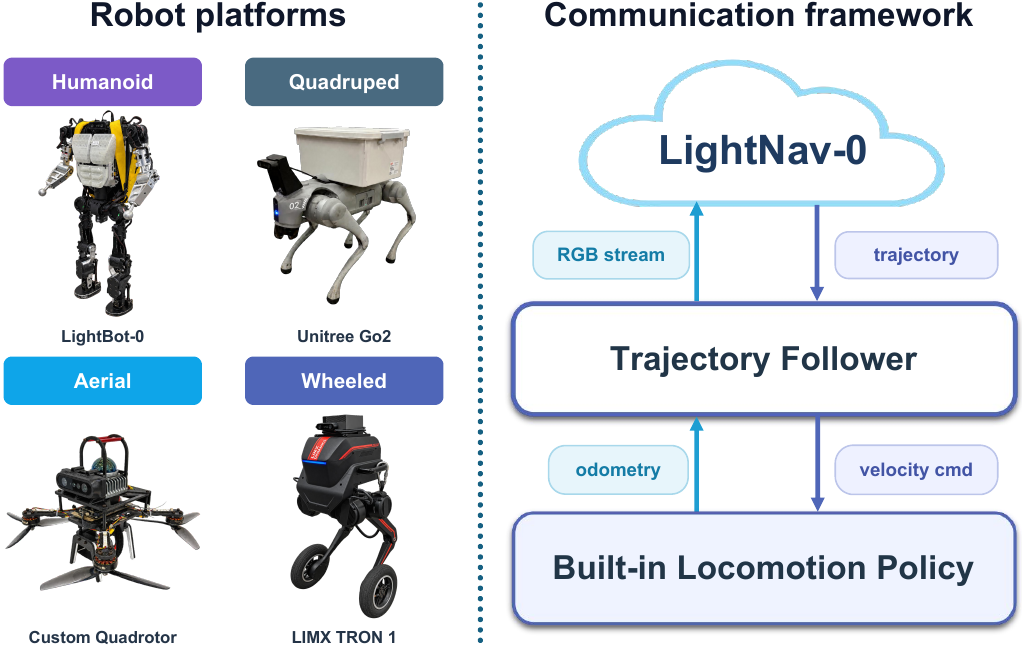}
    \caption{\textbf{Robot platforms and communication framework.} \ours receives RGB streams from humanoid, quadruped, aerial, and wheeled platforms and predicts trajectories that are executed by a shared trajectory follower interfacing with each robot's built-in locomotion policy through odometry and velocity commands.}
    \label{fig:robot-platform}
\end{figure}

We deploy \ours across four heterogeneous robot embodiments through the unified platform and communication interface shown in Fig.~\ref{fig:robot-platform}. This separation keeps the high-level RGB-to-trajectory policy unchanged, while a shared trajectory follower converts its predictions into the odometry and velocity commands required by each platform's built-in locomotion policy. The qualitative results therefore test transfer of a shared navigation model across both domains and embodiments, rather than separate policies tuned for individual robots.

\subsection{Ablation Study}

We ablate two components that connect spatial reasoning to embodied control: embodied-reasoning initialization and dual-channel pointing. Each variant is evaluated across the same 8 simulation settings used by the full model, covering instruction following, closed-vocabulary ObjectNav, and open-vocabulary ObjectNav.

\paragraph{Embodied-reasoning initialization}
Tab.~\ref{tab:ablation-er-init} compares policies initialized from the original Qwen3-VL checkpoint and from LightNav-ER. ER initialization raises the mean SR across the 8 settings from 60.5 to 63.0 (+2.5; 4.2\%) and the mean SPL from 38.8 to 39.9 (+1.1; 2.8\%). SR improves in all 8 settings, with the largest gains on MP3D (+4.2), OVON Unseen (+4.0), and HM3D v1 (+3.1). On HM3D v2, ER initialization increases SR from 75.6 to 77.2 (+1.6) and SPL from 40.3 to 41.5 (+1.2). The universal SR gains indicate that the spatial priors acquired during ER mid-training transfer consistently to goal-reaching reliability. The smaller and mixed SPL changes suggest that path efficiency remains more dependent on downstream navigation alignment.

\input{table/ablation_er_init}

\paragraph{Dual-channel pointing}
Removing affordance-point and object-point supervision degrades both SR and SPL on every benchmark in Tab.~\ref{tab:ablation-pointing}. Dual-channel pointing raises mean SR from 54.7 to 63.0 (+8.3; 15.2\%) and mean SPL from 34.3 to 39.9 (+5.6; 16.3\%). The largest improvements occur on HM3D v1, with +13.5 SR and +12.7 SPL; MP3D shows the next-largest SR gain (+12.0), while HM3D v2 shows the next-largest SPL gain (+8.1). The gains also persist across the seen, synonym, and unseen OVON splits, supporting pointing as a task-agnostic spatial interface rather than a cue specialized to instruction following or a fixed object taxonomy.

\input{table/ablation_pointing}

\subsection{Scaling Analysis}

\begin{figure*}[!t]
    \centering
    \includegraphics[width=\textwidth]{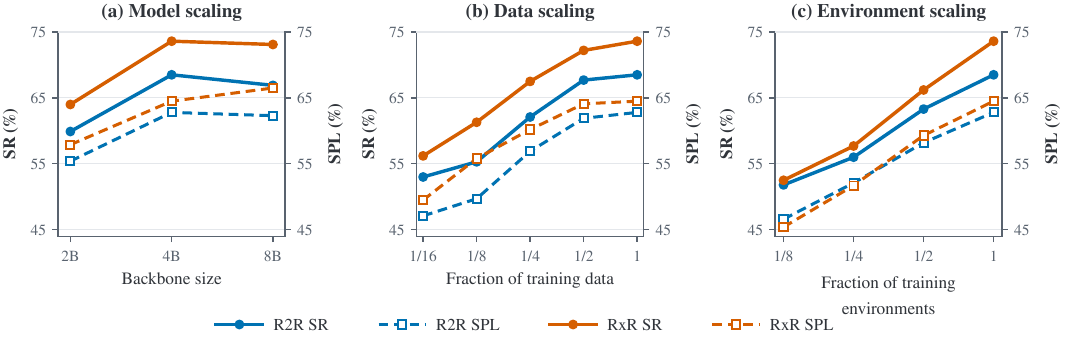}
    \caption{\textbf{Model, data, and environment scaling on continuous VLN.} The left and right vertical axes report SR and SPL, respectively. (a) Scaling the backbone from 2B to 4B parameters improves both metrics on R2R and RxR, whereas the 8B checkpoint produces mixed changes. (b) Increasing the fraction of training data yields monotonic gains with diminishing returns near the full-data regime. (c) Expanding the fraction of training environments consistently improves both metrics on both benchmarks.}
    \label{fig:scaling-analysis}
\end{figure*}

Fig.~\ref{fig:scaling-analysis} reveals distinct scaling behavior across model capacity, data volume, and environment coverage. Increasing the backbone from 2B to 4B raises R2R SR/SPL from 59.9/55.4 to 68.5/62.8 and RxR SR/SPL from 64.0/57.9 to 73.6/64.5, corresponding to gains of 6.6--9.6 points across the 4 measures. Scaling further to 8B is not consistently beneficial: R2R SR/SPL decrease by 1.6/0.5 points and RxR SR decreases by 0.5 points, although RxR SPL increases by 2.0 points. Among the tested checkpoints, the 4B model therefore provides the strongest capacity--performance trade-off; the mixed 8B result indicates that additional parameters alone do not guarantee better navigation performance at this scale.

Data scaling produces monotonic but saturating improvements. Moving from 1/16 to the full training set raises R2R SR/SPL from 53.0/47.1 to 68.5/62.8 and RxR SR/SPL from 56.2/49.5 to 73.6/64.5, yielding gains of 15.0--17.4 points. Most of this improvement is obtained before the final doubling: increasing the data fraction from 1/2 to the full set adds only 0.8/0.9 points on R2R and 1.4/0.4 points on RxR. Thus, additional training data remain beneficial over the tested range, but their marginal return diminishes as the training set approaches full scale.

Environment scaling is also monotonic and remains comparatively strong at matched data fractions. Increasing the training environments from $1/8$ to the full set improves R2R SR/SPL by 16.7/16.2 points and RxR SR/SPL by 21.1/19.1 points, with every intermediate increment improving all 4 measures. Over the matched $1/8$-to-full range, these gains exceed those from data scaling alone (13.2/13.1 points on R2R and 12.3/8.7 points on RxR). Within the tested sweeps, broader environment coverage is therefore the most reliable scaling axis, whereas model scaling saturates beyond 4B and data scaling shows diminishing returns near the full-data regime.

%% file: table/er_bench.tex
\begin{table*}[t]
    \centering
    \small
    \setlength{\tabcolsep}{2.6pt}
    \caption{\textbf{Embodied-reasoning and spatial-intelligence evaluation.} We compare 4B general-purpose VLMs, the 8B Molmo2-ER model, and our 4B LightNav-ER checkpoint on Point-Bench~\citep{cheng2025pointarena}, RefSpatial~\citep{zhou2025roborefer}, RoboSpatial~\citep{song2024robospatial}, Where2Place~\citep{yuan2024robopoint}, CV-Bench~\citep{tong2024cambrian}, ERQA~\citep{geminirobotics2025}, and EmbSpatial~\citep{du2024embspatial}. All values are percentages and higher is better. Avg. is the unweighted mean over all 8 benchmarks. \textbf{Bold} and \underline{underlined} denote best and second best.}
    \label{tab:er-bench}
    \begin{tabular}{lcccccccccc}
        \toprule
        Method & Params. & \shortstack{Point-\\Bench} & RefSpatial & \shortstack{RoboSpatial\\POI} & \shortstack{RoboSpatial\\VQA} & Where2Place & CV-Bench & ERQA & EmbSpatial & Avg. \\
        \midrule
        Qwen3-VL~\citep{bai2025qwen3vl} & 4B & 58.2 & 45.5 & \textbf{64.8} & 69.7 & \underline{64.0} & 85.6 & 39.5 & 77.6 & \underline{63.1} \\
        Qwen3.5-4B~\citep{qwen2026qwen35} & 4B & 60.4 & \underline{54.6} & 47.9 & 59.7 & 61.3 & 85.0 & 40.8 & 76.8 & 60.8 \\
        Molmo2-ER~\citep{fang2026molmoact2} & 8B & \textbf{77.3} & 52.5 & 32.0 & \textbf{73.4} & 54.0 & \underline{87.8} & \textbf{46.8} & \underline{78.8} & 62.8 \\
        \midrule
        \rowcolor{gray!20}
        \textbf{LightNav-ER} & 4B & \underline{64.5} & \textbf{57.4} & \underline{56.5} & \underline{71.9} & \textbf{76.6} & \textbf{88.4} & \underline{43.8} & \textbf{79.8} & \textbf{67.4} \\
        \bottomrule
    \end{tabular}
\end{table*}

%% file: table/vln_ce.tex
\begin{table*}[t]
    \centering
    \small
    \setlength{\tabcolsep}{2.1pt}
    \caption{\textbf{Performance on continuous vision-and-language navigation.} Comparison on the R2R~\citep{anderson2018vision} and RxR~\citep{ku2020room} val-unseen splits in continuous environments~\citep{Krantz2020BeyondTN}. S.RGB, Pano., Depth, and Odom. indicate single-view RGB, panoramic or multi-camera RGB, depth, and odometry inputs, respectively. $\ast$ denotes methods using a waypoint predictor. In monocular setting, \textbf{bold} and \underline{underlined} denote best and second best.}
    \label{tab:vln-ce}
    \begin{tabular}{lcccccccccccc}
        \toprule
        \multirow{2}{*}{Method} & \multicolumn{4}{c}{\textbf{Observation}} & \multicolumn{4}{c}{\textbf{R2R Val-Unseen}} & \multicolumn{4}{c}{\textbf{RxR Val-Unseen}} \\
        \cmidrule(lr){2-5} \cmidrule(lr){6-9} \cmidrule(lr){10-13}
        & S.RGB & Pano. & Depth & Odom. & NE$\downarrow$ & OS$\uparrow$ & SR$\uparrow$ & SPL$\uparrow$ & NE$\downarrow$ & SR$\uparrow$ & SPL$\uparrow$ & nDTW$\uparrow$ \\
        \midrule
        \multicolumn{13}{l}{\textit{Multi-view methods}} \\
        CMA$^\ast$~\citep{Krantz2020BeyondTN} & & $\checkmark$ & $\checkmark$ & $\checkmark$ & 6.20 & 52.0 & 41.0 & 36.0 & 8.76 & 26.5 & 22.1 & -- \\
        HPN+DN$^\ast$~\citep{krantz2021waypoint} & & $\checkmark$ & $\checkmark$ & $\checkmark$ & 6.31 & 40.0 & 36.0 & 34.0 & -- & -- & -- & -- \\
        Sim2Sim$^\ast$~\citep{krantz2022sim} & & $\checkmark$ & $\checkmark$ & $\checkmark$ & 6.07 & 52.0 & 43.0 & 36.0 & 8.76 & 26.5 & 22.1 & -- \\
        Reborn$^\ast$~\citep{an20221st} & & $\checkmark$ & $\checkmark$ & $\checkmark$ & 5.40 & 57.0 & 50.0 & 46.0 & 5.98 & 48.6 & 42.0 & -- \\
        GridMM$^\ast$~\citep{wang2023gridmm} & & $\checkmark$ & $\checkmark$ & $\checkmark$ & 5.11 & 61.0 & 49.0 & 41.0 & -- & -- & -- & -- \\
        DreamWalker$^\ast$~\citep{wang2023dreamwalker} & & $\checkmark$ & $\checkmark$ & $\checkmark$ & 5.53 & 59.0 & 49.0 & 44.0 & -- & -- & -- & -- \\
        ETPNav$^\ast$~\citep{an2024etpnav} & & $\checkmark$ & $\checkmark$ & $\checkmark$ & 4.71 & 65.0 & 57.0 & 49.0 & 5.64 & 54.7 & 44.8 & -- \\
        HNR$^\ast$~\citep{wang2024lookahead} & & $\checkmark$ & $\checkmark$ & $\checkmark$ & 4.42 & 67.0 & 61.0 & 51.0 & 5.50 & 56.3 & 46.7 & -- \\
        InstructNav~\citep{long2024instructnav} & & $\checkmark$ & $\checkmark$ & $\checkmark$ & 6.89 & -- & 31.0 & 24.0 & -- & -- & -- & -- \\
        AO-Planner~\citep{chen2024affordances} & & $\checkmark$ & $\checkmark$ & & 5.55 & 59.0 & 47.0 & 33.0 & -- & -- & -- & -- \\
        NavFoM~\citep{zhang2025embodied} & & $\checkmark$ & & & 4.61 & 72.1 & 61.7 & 55.3 & 4.74 & 64.4 & 56.2 & 65.8 \\
        NavForesee~\citep{liu2025navforesee} & & $\checkmark$ & & & 3.94 & 78.4 & 66.2 & 59.7 & 4.20 & 66.3 & 53.2 & -- \\
        SPAN-Nav~\citep{liu2026spannav} & & $\checkmark$ & & & 4.07 & 75.3 & 66.3 & 59.3 & 4.20 & 69.7 & 60.1 & 67.9 \\
        ABot-N0~\citep{chu2026abotn0} & & $\checkmark$ & & & 3.80 & 70.8 & 66.4 & 63.9 & 3.83 & 69.3 & 60.0 & -- \\
        Qwen-RobotNav-4B~\citep{zhang2026qwenrobotnav} & & $\checkmark$ & & & 3.80 & 77.2 & 69.5 & 63.6 & 3.80 & 75.2 & 65.0 & 71.9 \\
        Qwen-RobotNav-8B~\citep{zhang2026qwenrobotnav} & & $\checkmark$ & & & 3.53 & 78.5 & 72.1 & 66.6 & 3.58 & 76.5 & 65.7 & 72.5 \\
        ABot-N1~\citep{gong2026abotn1} & & $\checkmark$ & & & 3.91 & 71.7 & 68.3 & 66.6 & 3.43 & 70.9 & 61.4 & -- \\
        \midrule
        \multicolumn{13}{l}{\textit{Monocular methods}} \\
        NaVid~\citep{zhang2024navid} & $\checkmark$ & & & & 5.47 & 49.0 & 37.0 & 35.0 & -- & -- & -- & -- \\
        Uni-NaVid~\citep{zhang2024uni} & $\checkmark$ & & & & 5.58 & 53.5 & 47.0 & 42.7 & 6.24 & 48.7 & 40.9 & -- \\
        NaVILA~\citep{cheng2024navila} & $\checkmark$ & & & & 5.22 & 62.5 & 54.0 & 49.0 & 6.77 & 49.3 & 44.0 & -- \\
        StreamVLN~\citep{wei2025streamvln} & $\checkmark$ & & & & 4.98 & 64.2 & 56.9 & 51.9 & 6.22 & 52.9 & 46.0 & -- \\
        CorrectNav~\citep{yu2025correctnav} & $\checkmark$ & & & & 4.24 & 67.5 & 65.1 & \underline{62.3} & \underline{4.09} & 69.3 & 63.3 & -- \\
        DualVLN~\citep{wei2025dualvln} & $\checkmark$ & & & & \underline{4.05} & 70.7 & 64.3 & 58.5 & 4.58 & 61.4 & 51.8 & \textbf{70.0} \\
        InternVLA-N1~\citep{internvla2025} & $\checkmark$ & & $\checkmark$ & & 4.83 & 63.3 & 58.2 & 54.0 & 5.91 & 53.5 & 46.1 & 65.3 \\
        Qwen-VLA~\citep{wang2026qwenvla} & $\checkmark$ & & & & 5.10 & 69.0 & 57.3 & 51.2 & 5.80 & 59.6 & 47.8 & -- \\
        Qwen-RobotNav-4B~\citep{zhang2026qwenrobotnav} & $\checkmark$ & & & & 4.22 & \underline{73.6} & \underline{66.9} & 60.5 & 4.15 & 71.3 & 61.5 & 68.6 \\
        Qwen-RobotNav-8B~\citep{zhang2026qwenrobotnav} & $\checkmark$ & & & & 4.36 & 72.7 & 65.7 & 59.6 & 4.16 & \underline{73.4} & \underline{63.5} & \underline{69.9} \\
        \rowcolor{gray!20}
        \textbf{\ours} & $\checkmark$ & & & & \textbf{3.91} & \textbf{73.7} & \textbf{68.5} & \textbf{62.8} & \textbf{3.66} & \textbf{73.6} & \textbf{64.5} & 67.4 \\
        \bottomrule
    \end{tabular}
\end{table*}

%% file: table/objectnav.tex
\begin{table*}[t]
    \centering
    \small
    \setlength{\tabcolsep}{3.2pt}
    \caption{\textbf{Performance on object-goal navigation.} Comparison on MP3D~\citep{chang2017matterport3d} and HM3D~\citep{ramakrishnan2021hm3d} ObjectNav~\citep{batra2020objectnav}. In monocular setting, \textbf{bold} and \underline{underlined} denote best and second best.}
    \label{tab:objectnav}
    \begin{tabular}{lcccccccccc}
        \toprule
        \multirow{2}{*}{Method} & \multicolumn{4}{c}{\textbf{Observation}} & \multicolumn{2}{c}{\textbf{MP3D}} & \multicolumn{2}{c}{\textbf{HM3D v1}} & \multicolumn{2}{c}{\textbf{HM3D v2}} \\
        \cmidrule(lr){2-5} \cmidrule(lr){6-7} \cmidrule(lr){8-9} \cmidrule(lr){10-11}
        & S.RGB & Pano. & Depth & Odom. & SR$\uparrow$ & SPL$\uparrow$ & SR$\uparrow$ & SPL$\uparrow$ & SR$\uparrow$ & SPL$\uparrow$ \\
        \midrule
        \multicolumn{11}{l}{\textit{Multi-view methods}} \\
        WMNav~\citep{nie2025wmnav} & & $\checkmark$ & $\checkmark$ & $\checkmark$ & 45.4 & 17.2 & 58.1 & 31.2 & -- & -- \\
        Qwen-RobotNav-4B~\citep{zhang2026qwenrobotnav} & & $\checkmark$ & & & 52.2 & 16.0 & -- & -- & 75.6 & 30.6 \\
        Qwen-RobotNav-8B~\citep{zhang2026qwenrobotnav} & & $\checkmark$ & & & 48.8 & 17.7 & -- & -- & 71.2 & 33.0 \\
        \midrule
        \multicolumn{11}{l}{\textit{Monocular methods}} \\
        VLFM~\citep{yokoyama2024vlfm} & $\checkmark$ & & $\checkmark$ & $\checkmark$ & 36.4 & \underline{17.5} & 52.5 & 30.4 & 63.6 & 32.5 \\
        OpenFMNav~\citep{kuang2024openfmnav} & $\checkmark$ & & $\checkmark$ & $\checkmark$ & 37.2 & 15.7 & 52.5 & 24.1 & -- & -- \\
        SG-Nav~\citep{yin2024sg} & $\checkmark$ & & $\checkmark$ & $\checkmark$ & 40.2 & 16.0 & 54.0 & 24.9 & 49.6 & 25.5 \\
        TriHelper~\citep{zhang2024trihelper} & $\checkmark$ & & $\checkmark$ & $\checkmark$ & -- & -- & 56.5 & 25.3 & -- & -- \\
        FiLM-Nav~\citep{yokoyama2025film} & $\checkmark$ & & $\checkmark$ & $\checkmark$ & -- & -- & 61.7 & \underline{37.3} & \underline{77.0} & \underline{41.3} \\
        CogNav~\citep{cao2024cognav} & $\checkmark$ & & $\checkmark$ & $\checkmark$ & \underline{46.6} & 16.1 & 72.5 & 26.2 & -- & -- \\
        Uni-NaVid~\citep{zhang2024uni} & $\checkmark$ & & & & -- & -- & \underline{73.7} & 37.1 & -- & -- \\
        \rowcolor{gray!20}
        \textbf{\ours} & $\checkmark$ & & & & \textbf{53.3} & \textbf{21.2} & \textbf{74.5} & \textbf{43.9} & \textbf{77.2} & \textbf{41.5} \\
        \bottomrule
    \end{tabular}
\end{table*}

%% file: table/hm3d_ovon.tex
\begin{table*}[t]
    \centering
    \small
    \setlength{\tabcolsep}{3.2pt}
    \caption{\textbf{Performance on open-vocabulary object navigation.} Comparison on the HM3D-OVON benchmark~\citep{yokoyama2024hm3d} under seen-category, synonym, and unseen-category settings. In monocular setting, \textbf{bold} and \underline{underlined} denote best and second best.}
    \label{tab:hm3d-ovon}
    \begin{tabular}{lcccccccccc}
        \toprule
        \multirow{2}{*}{Method} & \multicolumn{4}{c}{\textbf{Observation}} & \multicolumn{2}{c}{\textbf{Seen}} & \multicolumn{2}{c}{\textbf{Synonyms}} & \multicolumn{2}{c}{\textbf{Unseen}} \\
        \cmidrule(lr){2-5} \cmidrule(lr){6-7} \cmidrule(lr){8-9} \cmidrule(lr){10-11}
        & S.RGB & Pano. & Depth & Odom. & SR$\uparrow$ & SPL$\uparrow$ & SR$\uparrow$ & SPL$\uparrow$ & SR$\uparrow$ & SPL$\uparrow$ \\
        \midrule
        \multicolumn{11}{l}{\textit{Multi-view methods}} \\
        NavFoM~\citep{zhang2025embodied} & & $\checkmark$ & & & 37.7 & 25.5 & 43.3 & 29.9 & 43.6 & 31.3 \\
        Qwen-RobotNav-4B~\citep{zhang2026qwenrobotnav} & & $\checkmark$ & & & 57.7 & 24.4 & 60.1 & 25.1 & 53.1 & 20.9 \\
        Qwen-RobotNav-8B~\citep{zhang2026qwenrobotnav} & & $\checkmark$ & & & 56.1 & 28.5 & 57.8 & 28.8 & 51.2 & 24.0 \\
        ABot-N0~\citep{chu2026abotn0} & & $\checkmark$ & & & 55.3 & 32.1 & 55.4 & 33.2 & 54.0 & 30.5 \\
        \midrule
        \multicolumn{11}{l}{\textit{Monocular methods}} \\
        VLFM~\citep{yokoyama2024vlfm} & $\checkmark$ & & $\checkmark$ & $\checkmark$ & 35.2 & 18.6 & 32.4 & 17.3 & 35.2 & 19.6 \\
        DAgRL+OD~\citep{yokoyama2024hm3d} & $\checkmark$ & & $\checkmark$ & $\checkmark$ & 38.5 & 21.1 & 39.0 & 21.4 & 37.1 & \underline{19.8} \\
        MTU3D~\citep{zhu2025mtu} & $\checkmark$ & & $\checkmark$ & $\checkmark$ & \underline{55.0} & \underline{23.6} & \underline{45.0} & 14.7 & \underline{40.8} & 12.1 \\
        Uni-NaVid~\citep{zhang2024uni} & $\checkmark$ & & & & 41.3 & 21.1 & 43.9 & \underline{21.8} & 39.5 & \underline{19.8} \\
        \rowcolor{gray!20}
        \textbf{\ours} & $\checkmark$ & & & & \textbf{55.3} & \textbf{31.2} & \textbf{54.6} & \textbf{29.6} & \textbf{47.0} & \textbf{24.2} \\
        \bottomrule
    \end{tabular}
\end{table*}

%% file: table/insight_bench_overall.tex
\begin{table}[t]
    \centering
    \small
    \setlength{\tabcolsep}{7pt}
    \caption{\textbf{Overall performance on INSIGHT-Bench.} \textbf{Bold} and \underline{underlined} denote best and second best.}
    \label{tab:insight-bench-overall}
    \begin{tabular}{lccc}
        \toprule
        Method & SR$\uparrow$ & SPL$\uparrow$ & NE$\downarrow$ (m) \\
        \midrule
        JanusVLN~\citep{zeng2025janusvln} & \underline{27.4} & \underline{24.0} & 4.89 \\
        NaVid~\citep{zhang2024navid} & 26.9 & 23.0 & \underline{4.25} \\
        Uni-NaVid~\citep{zhang2024uni} & 24.3 & 22.1 & 4.91 \\
        TAMP-Nav~\citep{feng2026tampnav} & 16.0 & 15.8 & 6.29 \\
        InternVLA-N1~\citep{internvla2025} & 11.7 & 11.0 & 5.45 \\
        StreamVLN~\citep{wei2025streamvln} & 11.6 & 10.8 & 6.56 \\
        \midrule
        \rowcolor{gray!20}
        \textbf{\ours} & \textbf{43.7} & \textbf{41.5} & \textbf{3.88} \\
        \bottomrule
    \end{tabular}
\end{table}

%% file: table/insight_bench_breakdown.tex
\begin{table*}[t]
    \centering
    \small
    \setlength{\tabcolsep}{2.05pt}
    \caption{\textbf{Fine-grained success rate on INSIGHT-Bench.} The first 5 columns group episodes by instruction type and the next 5 by scene type. \textbf{Bold} and \underline{underlined} denote best and second best.}
    \label{tab:insight-bench-breakdown}
    \begin{tabular}{lccccccccccc}
        \toprule
        \multirow{2}{*}{Method} & \multicolumn{5}{c}{\textbf{Instruction type}} & \multicolumn{5}{c}{\textbf{Scene type}} & \multirow{2}{*}{Avg.} \\
        \cmidrule(lr){2-6} \cmidrule(lr){7-11}
        & Base & Direction & Relation & Extremum & Ordinal & Apartment & House & Commercial & Institution & Outdoor & \\
        \midrule
        JanusVLN~\citep{zeng2025janusvln} & 32.5 & 28.5 & \underline{28.5} & 22.0 & \underline{25.8} & \underline{39.7} & 35.2 & \underline{26.7} & \underline{18.3} & \underline{16.7} & \underline{27.4} \\
        NaVid~\citep{zhang2024navid} & \underline{37.6} & \underline{29.7} & 18.7 & \underline{22.4} & 24.2 & 37.7 & \underline{39.8} & 24.6 & \underline{18.3} & 13.6 & 26.9 \\
        Uni-NaVid~\citep{zhang2024uni} & 32.9 & 29.3 & 21.8 & 16.8 & 19.7 & 39.3 & 28.7 & 23.1 & 15.5 & 14.0 & 24.3 \\
        TAMP-Nav~\citep{feng2026tampnav} & 18.1 & 14.2 & 17.6 & 10.8 & 20.8 & 26.8 & 18.5 & 15.9 & 9.6 & 8.3 & 16.0 \\
        InternVLA-N1~\citep{internvla2025} & 18.1 & 13.0 & 9.8 & 8.4 & 7.9 & 12.1 & 19.0 & 12.3 & 7.8 & 7.5 & 11.7 \\
        StreamVLN~\citep{wei2025streamvln} & 15.6 & 9.6 & 14.5 & 8.0 & 10.7 & 19.7 & 15.7 & 12.8 & 4.1 & 5.3 & 11.6 \\
        \midrule
        \rowcolor{gray!20}
        \textbf{\ours} & \textbf{45.1} & \textbf{57.7} & \textbf{37.8} & \textbf{37.2} & \textbf{38.2} & \textbf{61.1} & \textbf{50.5} & \textbf{42.1} & \textbf{29.2} & \textbf{34.2} & \textbf{43.7} \\
        \bottomrule
    \end{tabular}
    \vspace{10pt}

    \input{table/insight_bench_matrix}
\end{table*}

%% file: table/insight_bench_matrix.tex
\begingroup
\centering
\includegraphics[width=0.8\textwidth,pagebox=cropbox]{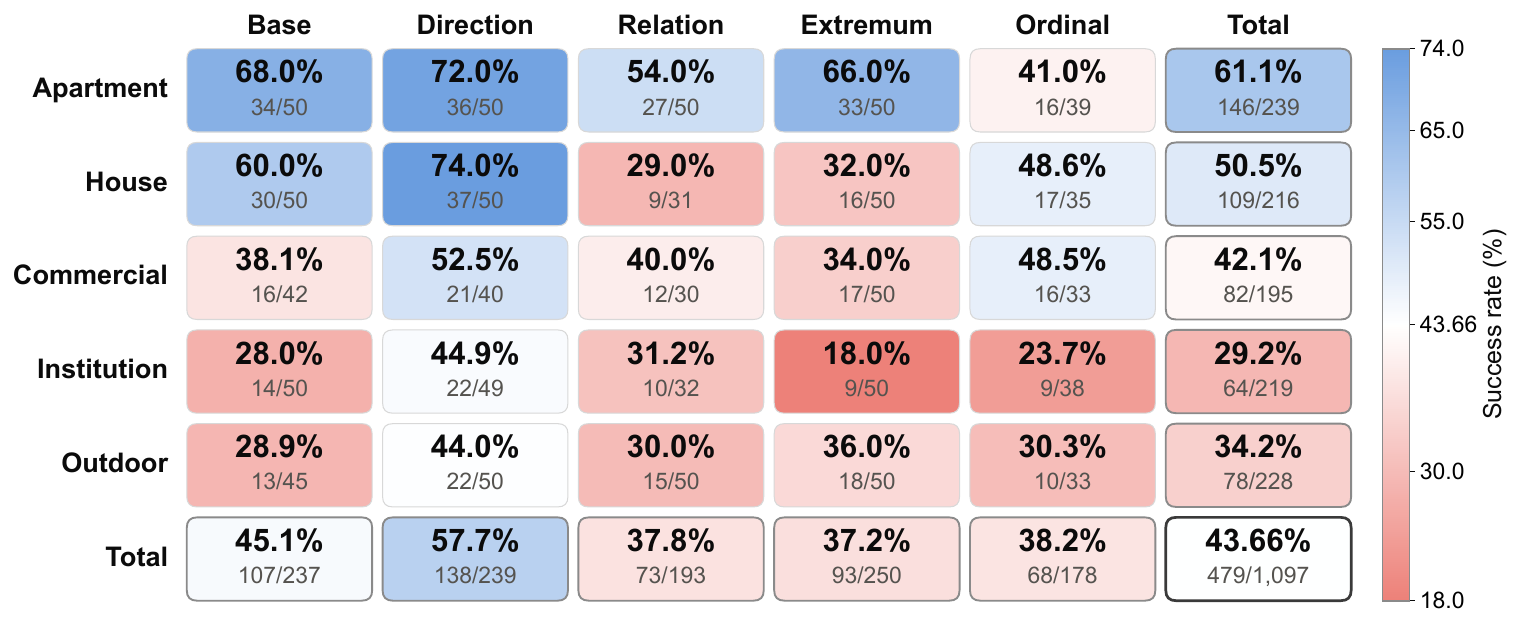}
\refstepcounter{figure}\label{fig:insight-bench-matrix}
\par\vspace{2pt}
\parbox{0.92\textwidth}{\small Fig.~\thefigure. \textbf{Joint scene--instruction capability matrix of \ours on the INSIGHT-Bench evaluation split.} Each cell reports SR and successful episodes over evaluated episodes. Warm cells fall below the overall SR of 43.66\%, whereas cool cells exceed it; saturation encodes the magnitude of this deviation on the colorbar scale, and thin borders mark row, column, and overall totals. The bottom-right cell is the aggregate result (479/1,097). The best and worst intersections are House--Direction (74.0\%, 37/50) and Institution--Extremum (18.0\%, 9/50), respectively.}
\endgroup

%% file: table/evt_bench.tex
\begin{table}[t]
    \centering
    \small
    \setlength{\tabcolsep}{2.9pt}
    \caption{\textbf{Performance on embodied visual tracking.} Comparison on EVT-Bench~\citep{wang2025trackvla}. In monocular setting, \textbf{bold} and \underline{underlined} denote best and second best.}
    \label{tab:evt-bench}
    \begin{tabular}{lcccccc}
        \toprule
        \multirow{2}{*}{Method} & \multicolumn{3}{c}{\textbf{STT}} & \multicolumn{3}{c}{\textbf{DT}} \\
        \cmidrule(lr){2-4} \cmidrule(lr){5-7}
        & SR$\uparrow$ & TR$\uparrow$ & CR$\downarrow$ & SR$\uparrow$ & TR$\uparrow$ & CR$\downarrow$ \\
        \midrule
        \multicolumn{7}{l}{\textit{Multi-view methods}} \\
        TrackVLA++~\citep{liu2025trackvla++} & 86.0 & 81.0 & 2.10 & 66.5 & 68.8 & 4.71 \\
        NavFoM~\citep{zhang2025embodied} & 86.0 & 80.5 & -- & 61.4 & 68.2 & -- \\
        CoMaTrack~\citep{liu2026comatrack} & 92.1 & 90.3 & 0.90 & 74.2 & 80.5 & 2.10 \\
        Qwen-RobotNav-4B~\citep{zhang2026qwenrobotnav} & 77.4 & 90.0 & 6.40 & -- & -- & -- \\
        Qwen-RobotNav-8B~\citep{zhang2026qwenrobotnav} & 78.6 & 89.7 & 5.70 & -- & -- & -- \\
        ABot-N0~\citep{chu2026abotn0} & 86.9 & 87.6 & 8.54 & 66.7 & 75.4 & 11.60 \\
        ABot-N1~\citep{gong2026abotn1} & 87.0 & 86.9 & 6.83 & 65.2 & 72.7 & 14.70 \\
        \midrule
        \multicolumn{7}{l}{\textit{Monocular methods}} \\
        TrackVLA~\citep{wang2025trackvla} & 85.1 & 78.6 & \underline{1.65} & 57.6 & 63.2 & 5.80 \\
        Uni-NaVid~\citep{zhang2024uni} & 53.3 & 67.2 & 12.60 & 31.9 & 50.1 & 21.30 \\
        VLingNav~\citep{wang2026vlingnav} & 88.4 & 81.2 & 2.07 & 67.6 & 73.5 & \underline{5.51} \\
        ReferTrack~\citep{ye2026refertrack} & \underline{89.4} & \textbf{92.5} & \textbf{1.60} & \underline{73.3} & \textbf{81.8} & 7.60 \\
        \rowcolor{gray!20}
        \textbf{\ours} & \textbf{91.7} & \underline{87.7} & 1.87 & \textbf{82.6} & \underline{80.1} & \textbf{4.62} \\
        \bottomrule
    \end{tabular}
\end{table}

%% file: table/ablation_er_init.tex
\begin{table*}[t]
    \centering
    \scriptsize
    \setlength{\tabcolsep}{3.2pt}
    \caption{\textbf{Effect of embodied-reasoning initialization.} We compare navigation policies initialized from the original Qwen3-VL checkpoint and from LightNav-ER across instruction following, closed-vocabulary ObjectNav, and open-vocabulary ObjectNav. All results are obtained with single-view RGB observations. \textbf{Bold} denotes the better result in each column.}
    \label{tab:ablation-er-init}
    \resizebox{\textwidth}{!}{%
    \begin{tabular}{lcccccccccccccccc}
        \toprule
        \multirow{2}{*}{Initialization} & \multicolumn{2}{c}{\textbf{R2R}} & \multicolumn{2}{c}{\textbf{RxR}} & \multicolumn{2}{c}{\textbf{HM3D v1}} & \multicolumn{2}{c}{\textbf{HM3D v2}} & \multicolumn{2}{c}{\textbf{MP3D}} & \multicolumn{2}{c}{\textbf{OVON Seen}} & \multicolumn{2}{c}{\textbf{OVON Synonyms}} & \multicolumn{2}{c}{\textbf{OVON Unseen}} \\
        \cmidrule(lr){2-3} \cmidrule(lr){4-5} \cmidrule(lr){6-7} \cmidrule(lr){8-9} \cmidrule(lr){10-11} \cmidrule(lr){12-13} \cmidrule(lr){14-15} \cmidrule(lr){16-17}
        & SR$\uparrow$ & SPL$\uparrow$ & SR$\uparrow$ & SPL$\uparrow$ & SR$\uparrow$ & SPL$\uparrow$ & SR$\uparrow$ & SPL$\uparrow$ & SR$\uparrow$ & SPL$\uparrow$ & SR$\uparrow$ & SPL$\uparrow$ & SR$\uparrow$ & SPL$\uparrow$ & SR$\uparrow$ & SPL$\uparrow$ \\
        \midrule
        Qwen3-VL & 65.8 & 59.9 & 72.6 & \textbf{64.9} & 71.4 & 42.0 & 75.6 & 40.3 & 49.1 & 19.8 & 53.7 & \textbf{31.2} & 52.5 & \textbf{29.7} & 43.0 & 22.4 \\
        LightNav-ER & \textbf{68.5} & \textbf{62.8} & \textbf{73.6} & 64.5 & \textbf{74.5} & \textbf{43.9} & \textbf{77.2} & \textbf{41.5} & \textbf{53.3} & \textbf{21.2} & \textbf{55.3} & \textbf{31.2} & \textbf{54.6} & 29.6 & \textbf{47.0} & \textbf{24.2} \\
        \bottomrule
    \end{tabular}
    }
\end{table*}

%% file: table/ablation_pointing.tex
\begin{table*}[t]
    \centering
    \scriptsize
    \setlength{\tabcolsep}{3.2pt}
    \caption{\textbf{Effect of dual-channel pointing.} We remove the affordance-point and object-point supervision while retaining the remaining model and action representation. All results are obtained with single-view RGB observations. \textbf{Bold} denotes the better result in each column.}
    \label{tab:ablation-pointing}
    \resizebox{\textwidth}{!}{%
    \begin{tabular}{lcccccccccccccccc}
        \toprule
        \multirow{2}{*}{Variant} & \multicolumn{2}{c}{\textbf{R2R}} & \multicolumn{2}{c}{\textbf{RxR}} & \multicolumn{2}{c}{\textbf{HM3D v1}} & \multicolumn{2}{c}{\textbf{HM3D v2}} & \multicolumn{2}{c}{\textbf{MP3D}} & \multicolumn{2}{c}{\textbf{OVON Seen}} & \multicolumn{2}{c}{\textbf{OVON Synonyms}} & \multicolumn{2}{c}{\textbf{OVON Unseen}} \\
        \cmidrule(lr){2-3} \cmidrule(lr){4-5} \cmidrule(lr){6-7} \cmidrule(lr){8-9} \cmidrule(lr){10-11} \cmidrule(lr){12-13} \cmidrule(lr){14-15} \cmidrule(lr){16-17}
        & SR$\uparrow$ & SPL$\uparrow$ & SR$\uparrow$ & SPL$\uparrow$ & SR$\uparrow$ & SPL$\uparrow$ & SR$\uparrow$ & SPL$\uparrow$ & SR$\uparrow$ & SPL$\uparrow$ & SR$\uparrow$ & SPL$\uparrow$ & SR$\uparrow$ & SPL$\uparrow$ & SR$\uparrow$ & SPL$\uparrow$ \\
        \midrule
        w/o pointing & 59.5 & 55.8 & 70.8 & 64.2 & 61.0 & 31.2 & 67.9 & 33.4 & 41.3 & 15.2 & 48.2 & 26.1 & 45.8 & 25.6 & 43.1 & 22.8 \\
        \textbf{\ours} & \textbf{68.5} & \textbf{62.8} & \textbf{73.6} & \textbf{64.5} & \textbf{74.5} & \textbf{43.9} & \textbf{77.2} & \textbf{41.5} & \textbf{53.3} & \textbf{21.2} & \textbf{55.3} & \textbf{31.2} & \textbf{54.6} & \textbf{29.6} & \textbf{47.0} & \textbf{24.2} \\
        \bottomrule
    \end{tabular}
    }
\end{table*}

%% file: section/conclusion.tex
\section{Conclusion}

We presented \ours, a compact generalist embodied navigation model that elicits the spatial intelligence of a pretrained VLM rather than introducing task-specific navigation architectures. Dual-channel pointing expresses task- and embodiment-agnostic spatial intent, while temporal history compression and hierarchical RVQ action tokens connect long-horizon visual context to precise short-horizon trajectories through the original autoregressive language-model head. Training on a unified corpus spanning $2\mathrm{K}{+}$ scenes and $4\mathrm{K}{+}$ hours of embodied trajectories aligns the same monocular RGB policy across instruction following, object goal navigation, and embodied visual tracking. LightNav-ER, the embodied-reasoning checkpoint from which \ours is initialized, attains the highest complete-set average across 8 embodied-reasoning benchmarks. The resulting \ours checkpoint achieves the strongest monocular success rates across all 10 public navigation simulation settings and transfers without task- or robot-specific model adaptation to four real-world robot embodiments. Together, these results indicate that compact VLMs can provide a unified and transferable substrate for generalist embodied navigation without relying on separate prediction heads for each task or platform.

Several directions can extend this framework. First, the current model uses a single decision pathway and does not explicitly separate high-frequency local control from slower semantic deliberation. A dual system could pair a lightweight reactive policy for obstacle avoidance and frequent trajectory correction with a slower VLM planner for long-horizon reasoning. Second, stronger pretraining on Internet-scale video could broaden open-world concept coverage and expose the model to rarer scenes, interactions, and motion patterns than curated embodied datasets alone.